\newif\ifRAL
\RALtrue 	%

\newif\ifTR
\TRfalse 	%

\newif\ifPrePrint
\PrePrinttrue

\newif\ifDraft
\Draftfalse

\ifRAL
\documentclass[letterpaper, 10 pt, journal, twoside]{IEEEtran}
\else
\documentclass[letterpaper, 10 pt, conference]{ieeeconf}
\fi

\IEEEoverridecommandlockouts

\ifRAL
\else
\fi		

\makeatletter

\let\proof\@undefined
\let\endproof\@undefined
\makeatother

\usepackage{amsmath} %
\usepackage{amssymb}  %
\usepackage{amsthm}
\usepackage{amsfonts}
\usepackage{mathtools}
\usepackage{acro}
\usepackage{bm}
\usepackage{cite}
\providecommand{\bm}{\pmb}

\theoremstyle{definition}

\theoremstyle{remark}

\usepackage{verbatim}

\usepackage{color}
\usepackage{subcaption}
\usepackage{graphicx}
\usepackage{caption}
\usepackage{siunitx}

\usepackage{optidef}

\usepackage{times}

\usepackage{xspace}

\usepackage[ruled,vlined,linesnumbered]{algorithm2e}
\usepackage[noend]{algorithmic}

\graphicspath{{images/}}

\usepackage[us]{datetime}
\usepackage{caption}
\usepackage[usenames,dvipsnames,table]{xcolor}

\usepackage{hyperref} %
\hypersetup{
    colorlinks,
    citecolor=black,
    filecolor=black,
    linkcolor=black,
    urlcolor=black,
    pdfauthor={},
    pdfsubject={},
    pdftitle={}
}
\usepackage[capitalise]{cleveref}
\crefformat{equation}{(#2#1#3)}
\let\labelindent\relax
\usepackage[inline]{enumitem}

\usepackage{todonotes}

\usepackage{graphicx}

\usepackage{latexsym}
\usepackage{color}

\usepackage{cite}

\usepackage{caption}

\usepackage{tikz}
\usetikzlibrary{shapes,arrows}
\usetikzlibrary{arrows.meta}
\usetikzlibrary{positioning,fit,backgrounds}
\usepackage{tabularx, booktabs}
\newcolumntype{Y}{>{\centering\arraybackslash}X}
\usepackage{multirow}
\usepackage{acro}
\usepackage{bm}
\usepackage{mathtools}
\usepackage{amsfonts}
\usepackage{amsmath}
\usepackage{subcaption}
\usepackage{xcolor}
\usepackage{siunitx}
\usepackage{hyperref}
\usepackage{svg}
\usepackage{optidef}

\DeclareAcronym{ASL}{short = ASL, long = Autonomous Systems Lab}
\DeclareAcronym{OMAV}{short = OMAV, long = Omnidirectional Micro Aerial Vehicle}
\DeclareAcronym{MAV}{short = MAV, long = Micro Aerial Vehicle}
\DeclareAcronym{DOF}{short = DOF, long = degrees of freedom}
\DeclareAcronym{PBC}{short = PBC, long = passivity-based control}
\DeclareAcronym{PH}{short = PH, long = Port-Hamiltonian}
\DeclareAcronym{NDT}{short = NDT, long = non-destructive testing}
\DeclareAcronym{PEMS}{short = PEMS, long = Power and Energy Monitoring System}
\DeclareAcronym{WTC}{short = WTC, long = wrench tracking controller}
\DeclareAcronym{PTC}{short = PTC, long = pose tracking controller}
\DeclareAcronym{MBE}{short = MBE, long = momentum-based wrench estimator}
\DeclareAcronym{ASIC}{short = ASIC, long = Axis-Selective Impedance Control}
\DeclareAcronym{MPC}{short = MPC, long = Model Predictive Control}
\DeclareAcronym{APhI}{short = APhI, long = Aerial Physical Interaction}
\DeclareAcronym{LLE}{short = LLE, long = Largest Lyapunov Exponent}
\DeclareAcronym{ICBF}{short = ICBF, long = Integral Control Barrier Function}
\DeclareAcronym{CBF}{short = CBF, long = Control Barrier Function}
\DeclareAcronym{COM}{short = CoM, long = Center of Mass}
\DeclareAcronym{HRI}{short = HRI, long = Human Robot Interaction}
\DeclareAcronym{DoF}{short = DoF, long = degree of freedom, short-plural = s, long-plural-form = degrees of freedom}
\DeclareAcronym{FT}{short = F/T, long = force and torque, short-indefinite = an, long-indefinite = a}

\newcommand{\EC}[1]{#1}
\newcommand{\rwThree}[1]{\orange{#1}}
\newcommand{\rwSix}[1]{\red{#1}}
\newcommand{\rwNine}[1]{\blue{#1}}
\newcommand{\rwEditor}[1]{\violet{#1}}

\renewcommand{\rwThree}[1]{#1}
\renewcommand{\rwSix}[1]{#1}
\renewcommand{\rwNine}[1]{#1}
\renewcommand{\rwEditor}[1]{#1}

\newcommand{\calC}{\mathcal{C}}
\newcommand{\calK}{\mathcal{K}}
\newcommand{\calL}{\mathcal{L}}
\newcommand{\PosSafeSet}{\mathcal{C}_{s,p}}
\newcommand{\ForceSafeSet}{\mathcal{C}_{s,\tau}}

\renewcommand{\vec}[1]{\bm{#1}}		
\renewcommand{\(}{\left(}		
\renewcommand{\)}{\right)}		
\renewcommand{\[}{\left[}		
\renewcommand{\]}{\right]}		
\newcommand{\matr}[1]{\bm{#1}}		
\newcommand{\blkdiag}[1]{\text{blkdiag}\left\{#1\right\}}		
\newcommand{\mat}[1]{\bm{#1}}		
\newcommand{\nR}[1]{\mathbb{R}^{#1}}		
\newcommand{\upperRomannumeral}[1]{\uppercase\expandafter{\romannumeral#1}}	

\newcommand{\vSpace}{\;\,}
\newcommand{\transpose}{^\top}
\newcommand{\refer}{^{\text{ref}}}
\newcommand{\W}[1]{\prescript{}{W}{#1}}
\newcommand{\B}[1]{\prescript{}{B}{#1}}

\newcommand{\maxEig}[1]{\lambda_{\text{max}}[#1]}

\renewcommand{\frame}[1]{\mathcal{F}_{#1}}		

\newcommand{\bfx}{\vec{x}}					
\newcommand{\bfU}{\mat{U}}					
\newcommand{\bfw}{\vec{w}}
\newcommand{\cbf}{h}
\newcommand{\powercbf}{\cbf_p}
\newcommand{\powercbfi}{\cbf_{p,i}}
\newcommand{\inputcbf}{\cbf_\tau}
\newcommand{\inputcbfi}{\cbf_{\tau,i}}
\newcommand{\pos}{\vec{p}}				
\newcommand{\vel}{\vec{v}}				
\newcommand{\veli}{v_i}				
\newcommand{\acc}{\dot{\vel}}				
\newcommand{\acci}{\dot{v}_i}

\newcommand{\rotMat}{\matr{R}}				
\newcommand{\eye}[1]{\matr{I}_{#1}}

\newcommand{\slack}{\vec{\delta}}
\newcommand{\slackPower}{\slack_p}
\newcommand{\slackInput}{\slack_\tau}
\newcommand{\slackWeight}{k_\delta}

\newcommand{\selectionMat}{\bm{\Lambda}}		

\newcommand{\frameW}{\frame{W}}			
\newcommand{\frameB}{\frame{B}}			

\newcommand{\rotMatWB}{\rotMat_{WB}}	
\newcommand{\angVel}{\vec{\omega}}

\DeclarePairedDelimiter{\norm}{\lVert}{\rVert} 

\newcommand{\wrench}{\bm{\tau}}

\newcommand{\bff}{\bm{f}}
\newcommand{\bfg}{\bm{g}}
\newcommand{\bfu}{\bm{u}}
\newcommand{\bfA}{\bm{A}}
\newcommand{\bfb}{\bm{b}}

\newcommand{\totalInertia}{\bm{M}}

\newcommand{\mass}{m}

\newcommand{\Coriolis}{\bm{C}}
\newcommand{\lle}{\hat{\lambda}^*}

\newcommand{\Klamb}{k_\lambda}
\newcommand{\KlambStrict}{k_\lambda'}
\newcommand{\Kspring}{\mat{K}_p}
\newcommand{\Kspringi}{k_{p,i}}
\newcommand{\Kpwalli}{k_{s,i}}
\newcommand{\Kspringigen}[1]{k_{p,#1}}
\newcommand{\Kforce}{\mat{K}_f}
\newcommand{\Kforcei}{k_{f,i}}
\newcommand{\KIforcei}{k_{I,i}}
\newcommand{\KIforce}{\mat{K}_I}

\newcommand{\Kdamping}{\mat{D}_v}
\newcommand{\Kdampingi}{d_{v,i}}
\newcommand{\Kdwalli}{d_{s,i}}
\newcommand{\Kdampingigen}[1]{d_{v,#1}}

\newcommand{\Kstiffness}{\bm{K}_s}
\newcommand{\Kjerk}{\bm{K}_\tau}

\newcommand{\jacobianPos}{\mat{U}_p}
\newcommand{\jacobianForce}{U_\tau}

\newcommand{\wrenchGravity}{\bm{g}}

\newcommand{\wrenchCommand}{\wrench_{c}}
\newcommand{\wrenchCommandDot}{\dot{\wrench}_{c}}
\newcommand{\wrenchCommandAugment}{\wrench_{a}}
\newcommand{\wrenchCommandAugmenti}{\tau_{a,i}}
\newcommand{\jerkCommand}{\dot{\wrench}_{a}}
\newcommand{\jerkCommandLimit}{\dot{\bar{\wrench}}_{a}}
\newcommand{\wrenchPosContr}{\wrench_{p}}
\newcommand{\wrenchPosContri}{\tau_{p,i}}
\newcommand{\wrenchForceContr}{\wrench_{f}}
\newcommand{\wrenchForceContri}{\tau_{f,i}}
\newcommand{\wrenchExt}{\wrench_\text{ext}}
\newcommand{\wrenchExti}{\tau_\text{ext,i}}
\newcommand{\wrenchExtMeas}{\wrench_\text{ext,meas}}

\newcommand{\wrenchCmd}{\wrench_{c}}

\newcommand{\error}{\bm{e}}
\newcommand{\errorPos}{\bm{e}_{p}}
\newcommand{\errorAng}{\bm{e}_{R}}
\newcommand{\errorVel}{\bm{e}_{v}}
\newcommand{\errorAngVel}{\bm{e}_{\omega}}

\newcommand{\errorVeli}{e_{v,i}}

\newcommand{\errorPosStacked}{\tilde{\error}_p}
\newcommand{\errorPosWallStacked}{\tilde{\error}_s}
\newcommand{\errorPosWallStackedi}{\tilde{e}_{s,i}}
\newcommand{\errorVelStacked}{\tilde{\error}_v}
\newcommand{\errorAccStacked}{\dot{\tilde{\error}}_v}
\newcommand{\errorPosStackedi}{\tilde{e}_{p,i}}
\newcommand{\errorVelStackedi}{\tilde{e}_{v,i}}
\newcommand{\errorVelStackedOne}{\tilde{e}_{v,1}}
\newcommand{\errorVelStackedSix}{\tilde{e}_{v,6}}
\newcommand{\errorAccStackedi}{\dot{\tilde{e}}_{v,i}}

\newcommand{\errorWrench}{\dot{\error}_{\tau}}
\newcommand{\errorWrenchi}{\dot{e}_{\tau,i}}
\newcommand{\errorWrenchIntegral}{\error_{\tau}}
\newcommand{\errorWrenchIntegrali}{e_{\tau,i}}

\newcommand{\pLim}{\bar{p}}

\newcommand{\powercbfGain}{\gamma_p}
\newcommand{\inputcbfGain}{\gamma_\tau}

\ifRAL %
\fi

\author{Eugenio Cuniato, Nicholas Lawrance, Marco Tognon, Roland Siegwart%
	\ifRAL
		\thanks{Manuscript received: 24,\,02,\,22; Revised 05,\,05,\,22 ; Accepted 05,\,05,\,22.}%
		\thanks{This paper was recommended for publication by Editor Pauline Pounds upon evaluation of the Associate Editor and Reviewers' comments. } %
	\fi
	\thanks{All authors are with the Autonomous Systems Lab (ASL), ETH Zurich. Corresponding author: {\tt \footnotesize \href{mailto:ecuniato@ethz.ch}{ecuniato@ethz.ch}.}}%
	\thanks{The research leading to this results has been supported by the AERO-TRAIN project, European Union's Horizon 2020 research and innovation programme under the Marie Skłodowska-Curie grant agreement No 953454 and by armasuisse S+T, NCCR Robotics, and NCCR Digital Fabrication. The authors are solely responsible for its content.}
	\ifRAL
		\thanks{Digital Object Identifier (DOI): see top of this page.}
	\fi
}

\ifRAL %
\title{Power-based Safety Layer for Aerial Vehicles in Physical Interaction using Lyapunov Exponents}

\else %
\title{\bf Power-based Safety Layer for Aerial Vehicles in Physical Interaction using Lyapunov Exponents}
\fi

\ifPrePrint
\makeatletter
\def\ps@titlepagestyle{
	\def\@oddfoot{}\def\@evenfoot{}
	\def\@oddhead{\textcolor{red}{\sf\footnotesize Preprint version, final version at http://ieeexplore.ieee.org/ \hfill IEEE Robotics and Automation Letters 2022}}
	\def\@evenhead{\textcolor{red}{\sf\footnotesize  Preprint version, final version at http://ieeexplore.ieee.org/  \hfill IEEE Robotics and Automation Letters 2022}}%
}%
\makeatother
\makeatletter
\def\ps@headings{
	\def\@oddfoot{\textcolor{red}{\sf\footnotesize  Preprint version, final version at http://ieeexplore.ieee.org/ \hfill \thepage \;\;~\hfill~\hfill IEEE Robotics and Automation Letters 2022}}\def\@evenfoot{\hfill\thepage\hfill}
	\def\@oddhead{}\def\@evenhead{}%
}%
\makeatother
\fi	
\ifDraft
\makeatletter
\def\ps@titlepagestyle{
	\def\@oddfoot{}\def\@evenfoot{}
	\def\@oddhead{\textcolor{red}{\sf Draft version  \hfill Confidential}}
	\def\@evenhead{\textcolor{red}{\sf  Draft version  \hfill Confidential}}%
}%
\makeatother
\makeatletter
\def\ps@headings{
	\def\@oddfoot{\textcolor{red}{\sf  Draft version  \hfill Confidential}}\def\@evenfoot{\hfill\thepage\hfill}
	\def\@oddhead{}\def\@evenhead{}%
}%
\makeatother
\usepackage{draftwatermark}
\SetWatermarkText{Confidential draft}
\SetWatermarkScale{0.6}
\SetWatermarkLightness{0.9} 

\fi	
\begin{document}

\maketitle

\begin{abstract}
	As the performance of autonomous systems increases, safety concerns arise, especially when operating in non-structured environments.
	To deal with these concerns, this work presents a  safety layer for mechanical systems that detects and responds to unstable dynamics caused by external disturbances.
	The safety layer is implemented independently and on top of already present nominal controllers, like pose or wrench tracking, and limits power flow when the system's response would lead to instability.
	This approach is based on the computation of the \ac{LLE} of the system's error dynamics, which represent a measure of the dynamics' divergence or convergence rate.
	By actively computing this metric, divergent and possibly dangerous system behaviors can be promptly detected.
	The \ac{LLE} is then used in combination with \acp{CBF} to impose power limit constraints on a jerk controlled system.
	The proposed architecture is experimentally validated on an \ac{OMAV} both in free flight and interaction tasks.
\end{abstract}
\ifRAL %
	\else %
	{} %
\fi

\section{Introduction}

\begin{figure}[t]
	\centering
	\includegraphics[width=\columnwidth]{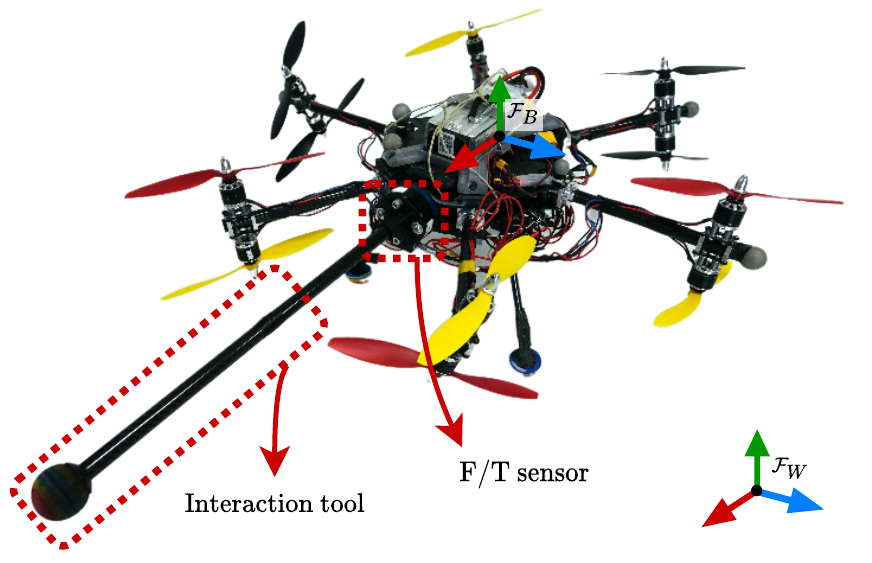}
	\caption{Picture of the \ac{OMAV} aerial robot equipped with a rigid interaction tool.}
	\label{fig:omav}
\end{figure}

Aerial robotics is a promising research field with an increasing number of applications as well as innovative aerial systems~\cite{Ollero, RuggieroReview}.
Among them, the development of fully actuated \acp{MAV} has grown rapidly in recent years~\cite{8485627, Tognon2019, Ryll20196DIC}.
A specific class of these are tiltrotor \acp{OMAV} (see Fig.~\ref{fig:omav}), capable of generating a six \acp{DoF} control wrench independently of their attitude.
Their actuation capability has proved fundamental for obtaining high disturbance rejection and performance in flight, \EC{ while also allowing physical interaction tasks that might be otherwise impossible for standard multirotors}~\cite{9295362,Brunner2020}.

As these platforms are employed in more and more tasks, safety concerns start arising, especially during \ac{APhI}.
\EC{In case of unexpected disturbances, \rwNine{some} controllers might lead to instability or damages of the platform if the danger is not detected and addressed in time.}
\rwNine{A possibility in this case would be resorting to robust control techniques with nonlinear disturbance observers for trajectory tracking \cite{Laing} or complex interaction \cite{LeeDOB} tasks.}
\rwNine{Another well-known approach in the literature} employs passivity-based control laws, that are notably robust against disturbances~\cite{Yuksel2019,Coelho2021}.
In this scenario, energy tanks have been successfully exploited to increase the robustness of the overall control algorithm~\cite{Rashad2019,7180405}.
They can be used as a versatile safety layer on top of non-passive nominal controllers to enforce the passivity of the tank-augmented dynamics.
In the energy tank theory, the passivity constraint acts to limit the system's energy, but only implicitly the power.
This requires the definition of an amount of energy that the tank can allow before entering into a strict passivity regime.
Some insights and interesting approaches on how to choose this energy budget can be found in the literature~\cite{7139036, rashad2021energy}.
However, how to generally approach this problem remains an open question.
An insufficient energy budget can lead to a premature interruption of the task, whereas excess energy no longer ensures safety.

The main goal of this paper is to overcome the limitation of energy budgets while still retaining the versatility of the safety layer approach.
To accomplish this goal, two elements are necessary: \textit{(i)} to impose constraints directly on the power level; and \textit{(ii)} a metric that can change the power limit according to a \textit{danger level}.
Regarding the first point, \acp{CBF} have been widely used to enforce safety constraints on mechanical systems~\cite{Ames2019}.
This approach has already been employed to enforce energy tank constraints in integral form~\cite{Benzi2021}, as well as power passivity constraints in~\cite{Ames2021}.
This last work constantly enforces a passive constraint (zero power flow limit) in all cases, without allowing for positive or negative limits.
In our framework we allow instead for the full spectrum of power limits, allowing for positive limits when the system is safely accomplishing the task, or imposing dissipation when some unexpected dangerous event happens \rwSix{(sudden impacts, changes in the environment, loss of contact during interaction, etc.)}.

\rwNine{Power safety policies were already employed inside an energy tank framework in \cite{brunner2022energy}. In this work, different power flow adaptation laws were presented specifically tailored for the proposed interaction task.
	In our work we employ a finite-time computation of the \ac{LLE} of the error dynamics~\cite{Dabrowski2011} as a metric for power adaptation, without the necessity of the energy tank framework.}
The computation of Lyapunov exponents, and in particular of the \ac{LLE}, gives a measure of convergence of adjacent state trajectories, showing possible unstable or chaotic behaviors. For example, \ac{LLE} has been used in~\cite{Mccue2011} to predict vessel capsizes.
\subsection*{Contributions}
In this work we design a power flow safety layer with an \ac{LLE}-based metric.
The \ac{LLE} metric measures the closed-loop system convergence rate and can be used to determine when the system is showing a divergent behavior.
This allows the system to be passive and even enforce dissipation only when divergence is detected, otherwise allowing the system to complete the commanded task.
\EC{
	Importantly, we analytically characterize the constraints imposed on the system dynamics by the chosen safety policy, showing that the control objectives remain reachable.}
Moreover, the proposed approach does not require the definition of energy budgets, a common limitation with existing energy-tank based approaches, which can lead to premature interruption of the desired task.
In the end, the safety layer is experimentally validated on both pose and wrench tracking tasks with an \ac{OMAV}.
\rwSix{In particular, we focus on the interaction with moving objects, like a cart pushing task.}
Notice that we propose this framework for aerial robots because for this class of systems safety is of particular importance.
However, this safety layer can be generally applied to any fully actuated mechanical system.

\begin{figure*}[t]
	\centering
	\centerline{\includegraphics[width=\textwidth]{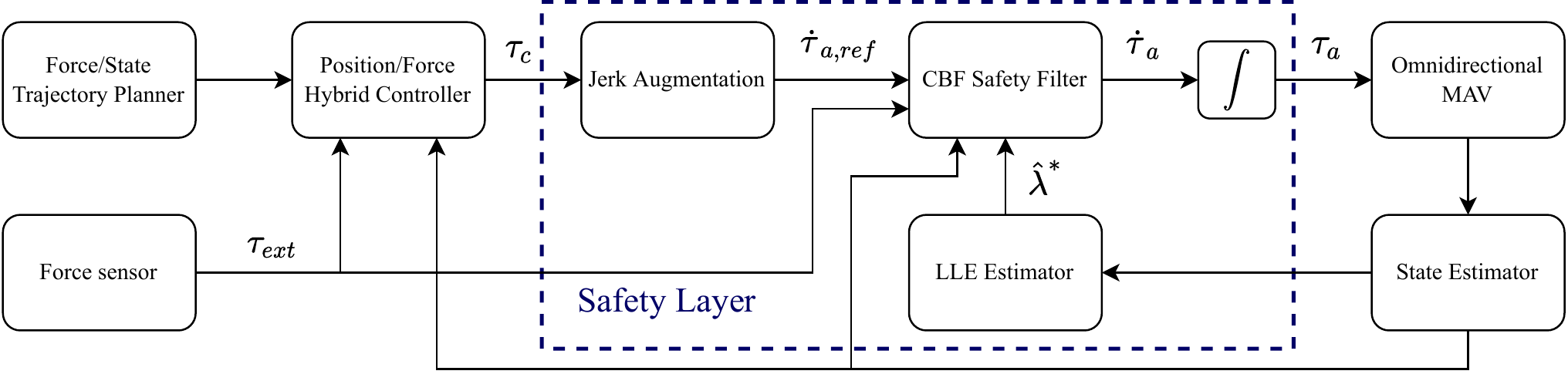}}
	\caption{\ac{OMAV} control scheme. The dashed line contains the proposed safety layer.}
	\label{fig:control_scheme}
\end{figure*}
\section{Model and Control}
\EC{The aerial robot considered in this work is an \ac{OMAV} composed of a rigid body with six independently tiltable rotor arms (Fig.~\ref{fig:omav}).
	Each arm has two coaxially-aligned propellers, for a total of twelve propellers.
	By controlling both the arm tilt angles and the propellers' speed, the platform is fully actuated in any orientation, i.e. it can always exert a full 6-\ac{DoF} wrench.
}

\EC{Consider a fixed world frame $\frameW$ and a body frame $\frameB$ attached to the drone's body located in its \ac{COM}.
	We describe the position and rotation of $\frameB$ with respect to $\frameW$ by the vector $\W{\pos} \in \nR{3}$ and the rotation matrix $\rotMatWB \in \nR{3 \times 3}$, respectively.}
Then, as already in~\cite{9295362,Brunner2020}, the Lagrangian model of the \ac{OMAV} can be expressed as~\footnote{This general dynamic formulation can be obtained for any fully actuated mechanical system.}
\begin{equation}
	\label{omav_original_model}
	\totalInertia \B\acc + \Coriolis(\B\angVel) + \wrenchGravity(\rotMatWB) = \wrenchCommand' + \wrenchExt ,
\end{equation}
with inertia matrix $\totalInertia \in \nR{6\times 6}$, control wrench $\wrenchCommand' \in \nR{6}$, external wrench $\wrenchExt\in \nR{6}$, gravity wrench $\wrenchGravity(\rotMatWB) \in \nR{6}$, centrifugal and Coriolis wrench $\Coriolis(\B\angVel) \in \nR{6}$ and $\B\vel=\left[ \B\vel_p\transpose \vSpace \B\angVel\transpose\right]\transpose \in \nR{6}$ system's twist all expressed in the body frame.
\rwNine{Specifically, $\totalInertia$ is expressed w.r.t. the body frame and thus constant. Moreover, by assuming the \ac{OMAV}'s body to be rigid and the body axes corresponding with the principal axes of inertia, we can consider $\totalInertia$ to be diagonal as well.}

Since the system is fully actuated, we can apply the feedback linearizing input $\wrenchCommand' = \Coriolis(\B\angVel) + \wrenchGravity(\rotMatWB) + \wrenchCommand$. The dynamics of the system \eqref{omav_original_model} becomes
\begin{equation}
	\label{omav_fb}
	\totalInertia \B\acc = \wrenchCommand + \wrenchExt .
\end{equation}
\rwSix{Notice that this compensation only requires knowledge of the system's angular velocity and attitude. In our tests these are obtained from a motion capture system, but outdoor scenarios would require on-board state estimation using IMU and/or optical sensors.}
Through the control input $\wrenchCommand$, either a pose tracking control action $\wrenchPosContr$ or \iac{WTC} $\wrenchForceContr$ can be implemented as
\begin{equation}
	\label{selected_control}
	\wrenchCommand = (\eye{6} - \selectionMat)\wrenchPosContr + \selectionMat\wrenchForceContr ,
\end{equation}
where $\selectionMat$ is a  $(6\times6)$ diagonal binary selection matrix.
\EC{Notice that the proposed safety framework is not limited to the choice of a hybrid position/force controller. In case of different controllers, only the \acp{LLE} computation in section \ref{LLE_computation} should be modified according to the new control errors.}
\rwSix{In the end, the control wrench $\wrenchCommand'$ is mapped into the \ac{OMAV}'s arm tilt angles and rotor speeds as in \cite{9295362}, assuming a quadratic propulsion model for the propeller's thrust.}

\subsection{Pose tracking control}
\EC{The pose tracking controller allows the system to follow a commanded reference state trajectory $(\cdot)\refer$ defined in terms of position and attitude references, as well as their derivatives.}
The control action is then computed as
\begin{equation}
	\label{position_control}
	\wrenchPosContr = \totalInertia \B\acc\refer + \Kdamping \errorVelStacked + \Kspring\errorPosStacked ,
\end{equation}
with $\Kdamping \in \nR{6\times 6}$ and $\Kspring \in \nR{6\times 6}$ positive definite matrices of damping and stiffness respectively. We write the stacked errors for the position and attitude dynamics as $\errorVelStacked = \left[ \errorVel\transpose \errorAngVel\transpose \right]\transpose$ and $\errorPosStacked = \left[ \errorPos\transpose \errorAng\transpose \right]\transpose $, where
\begin{subequations}
	\label{position_errors}
	\begin{IEEEeqnarray} {ll}
		\errorPos & = \rotMatWB \( \W{\pos} - \W{\pos}\refer \)\\
		\errorVel & = \B\vel_p - \rotMatWB \W\vel_p\refer \\
		\errorAng & = \frac{1}{2} \( \rotMatWB\refer{}\transpose \rotMatWB - \rotMatWB\transpose \rotMatWB\refer \)^\vee\\
		\errorAngVel & = \B\angVel - \rotMatWB \W\angVel\refer ,
	\end{IEEEeqnarray}
\end{subequations}
with the \textit{vee}-operator $\star^\vee$ used to extract a vector from a skew-symmetric matrix.
Applying the control action in \eqref{position_control} to the system in \eqref{omav_fb}, yields the closed loop dynamics
\begin{equation}
	\label{impedance_dynamics}
	\totalInertia \errorAccStacked + \Kdamping \errorVelStacked + \Kspring\errorPosStacked = \wrenchExt .
\end{equation}
Assume the matrices $\totalInertia = \text{diag}\{\mass_1,\dots,\mass_6\} $, $\Kdamping= \text{diag}\{\Kdampingigen{1},\dots,\Kdampingigen{6}\}$ and $\Kstiffness=  \text{diag}\{\Kspringigen{1},\dots,\Kspringigen{6}\}$ to be diagonal.
Then the closed loop dynamics can be decoupled along the six spatial \acp{DoF} and written as
\begin{equation}
	\label{impedance_dynamics_i}
	\mass_i \errorAccStackedi + \Kdampingi \errorVelStackedi + \Kspringi\errorPosStackedi = \wrenchExti , \forall i \in \{1,  \dots, 6\}.
\end{equation}

\subsection{Wrench tracking control}
The wrench tracking control action is computed as
\rwNine{
	\begin{equation}
		\label{wrench_control}
		\wrenchForceContr = -\wrenchExt\refer + \Kforce(\wrenchExtMeas - \wrenchExt\refer) + \KIforce \int (\wrenchExtMeas - \wrenchExt\refer),
	\end{equation}
	with reference wrench trajectory $\wrenchExt\refer \in \nR{6}$, $\Kforce$,$\KIforce \in \nR{6\times 6}$ diagonal and positive definite matrices, and $\wrenchExtMeas \in \nR{6}$ the wrench acting on the \ac{OMAV} body measured with the \ac{FT} sensor.
	With a slight abuse of notation, we define for later convenience the wrench control error as $\errorWrench = \wrenchExtMeas - \wrenchExt\refer$, and its integral error as $\errorWrenchIntegral = \int (\wrenchExtMeas - \wrenchExt\refer)dt$.
}
When the control action \eqref{wrench_control} is applied to the system in \eqref{omav_fb}, one gets the closed loop dynamics
\begin{equation}
	\label{force_dynamics}
	\left(\Kforce + \eye{6}\right)\errorWrench + \KIforce\errorWrenchIntegral = \totalInertia \B\acc.
\end{equation}
Since $\totalInertia$, $\Kforce$ and $\KIforce$ are diagonal, the closed loop dynamics can be again decoupled along the six spatial \acp{DoF} as
\begin{equation}
	\label{force_dynamics_i}
	\left(\Kforcei + 1\right)\errorWrenchi + \KIforcei\errorWrenchIntegrali = \mass_i \B\acci , \forall i \in \{1, \dots, 6\}.
\end{equation}

\section{Safety layer}
This section presents the power flow-based safety layer.
\EC{We first show how to compute the \ac{LLE} on the closed loop error dynamics.
	Then we present the CBF-based optimization problem used to control the system's power flow.
	Finally, a \ac{LLE}-based power adaptation law is designed and enforced with the CBF framework.}
The safety layer is schematically represented in Fig.~\ref{fig:control_scheme}, together with overall system architecture.

\subsection{\ac{LLE} computation}
\label{LLE_computation}
Consider a linear (or locally linearized) autonomous dynamic system
\begin{equation}
	\label{lyap_general_system}
	\dot{\bfx} = \bfU \bfx ,
\end{equation}
with system's state $\bfx \in \nR{n}$ and Jacobian matrix $\bfU \in \nR{n\times n}$.
\rwNine{We assume $\bfU$ to be diagonalizable with $n$ distinct eigenvalues.}
These dynamics can be projected along the system's eigenvectors $\bfw_i \in \nR{n}$ to obtain the eigenvalues $\lambda_i \in \nR{}$ as
\begin{equation}
	\dot{x}_i = \lambda_i x_i ,
\end{equation}
where $x_i =\bfx \cdot \bfw_i$ is the component of the vector $\bfx$ in the direction of $\bfw_i$, with $i = 1\dots n$.
As it is well known from the theory of linear systems, all these single evolutions contribute together to the system's modal decomposition
\begin{equation}
	\label{modal_decomposition}
	\bfx(t) = c_1 \bfw_1 e^{\lambda_1 t} + c_2 \bfw_2 e^{\lambda_2 t} + \dots + c_n \bfw_n e^{\lambda_n t} ,
\end{equation}
where the coefficients $c_1, \dots , c_n$ depend on the system's initial conditions.
Considering $\lambda^*$ as the biggest eigenvalue, the other contributions in \eqref{modal_decomposition} can be assumed negligible as $t \rightarrow \infty$, with the dynamics evolving according to $\bfx = c^* \bfw^* e^{\lambda^* t}$. We refer to $\lambda^*$ as the \ac{LLE}, computed with
\begin{equation}
	\label{LLE}
	\lambda^* = \frac{\bfx^\top \dot{\bfx}}{\|\bfx\|^2} .
\end{equation}
\EC{In practice, this punctual computation of $\lambda^*$ will be affected by both noise and oscillatory behaviors of the dynamics, that will not be perfectly exponential \cite{Dabrowski2011}.
	We then estimate the \ac{LLE} by low-pass filtering the quantity in \eqref{LLE}, obtaining $\lle$.}
This value determines if the system is currently asymptotically stable $\lle < 0$ or unstable $\lle > 0$.
Notice that this quantity is well defined also when asymptotically stable dynamics approach the equilibrium in $\left(\dot{\bfx}, \bfx\right) = \left(0, 0\right)$.
In that case, it is
\begin{equation}
	\label{eq:lle_nominal}
	\lim_{\left(\dot{\bfx}, \bfx\right) \to \left(0, 0\right)} \lambda^* = \maxEig{U} ,
\end{equation}
with $\maxEig{U}$ the biggest real part of the eigenvalues of the Jacobian matrix $\bfU$. The \ac{LLE} estimation for both pose and wrench controllers can be now derived as follows.

\subsubsection{Pose tracking \ac{LLE}}
\EC{
	To compute the \ac{LLE}, we first transform the closed loop error dynamics in \eqref{impedance_dynamics_i} as \eqref{lyap_general_system}
	\begin{equation}
		\label{impedance_dynamics_matrix}
		\begin{bmatrix}
			\errorVelStackedi \\
			\errorAccStackedi
		\end{bmatrix}
		=
		\begin{bmatrix}
			0                          & 1                           \\
			-\frac{\Kspringi}{\mass_i} & -\frac{\Kdampingi}{\mass_i}
		\end{bmatrix}
		\begin{bmatrix}
			\errorPosStackedi \\
			\errorVelStackedi
		\end{bmatrix}
		+\frac{1}{\mass_i}
		\begin{bmatrix}
			0 \\
			\wrenchExti
		\end{bmatrix} ,
	\end{equation}
	where the Jacobian matrix\footnote{\rwNine{This matrix is always diagonalizable, unless $\Kdampingi = 2\sqrt{\Kspringi\mass_i}$.}} is represented by
	\begin{equation}
		\label{position_jacobian}
		\jacobianPos
		=
		\begin{bmatrix}
			0                          & 1                           \\
			-\frac{\Kspringi}{\mass_i} & -\frac{\Kdampingi}{\mass_i}
		\end{bmatrix}.
	\end{equation}
}
Then, each pose tracking \ac{LLE} $\lle_{p,i}$ can be computed for each second-order dynamics in \eqref{impedance_dynamics_i} as
\begin{equation}
	\label{impedance_lle}
	\lambda^*_{p,i} = \frac{ \[\errorVelStackedi \vSpace \errorPosStackedi\] \[\errorAccStackedi  \vSpace \errorVelStackedi\]\transpose }{\|\[\errorVelStackedi  \vSpace \errorPosStackedi\]\|^2} , \forall i \in \lbrace 1, \dots, 6\rbrace.
\end{equation}
Notice that theoretically the acceleration $\errorAccStackedi$ is necessary to compute the previous quantity.
\EC{
	For instance, it might be obtained by numerically filtering the velocity.
	However, we will show in Section~\ref{Par:safe_sets} that the proposed method enforces safety even when assuming $\errorAccStackedi=0$, as we do.
}

\subsubsection{Wrench tracking \ac{LLE}}
\EC{
	Similarly to \eqref{impedance_dynamics_matrix}, we can rewrite the wrench control dynamics as
	\begin{equation}
		\label{force_dynamics_matrix}
		\errorWrenchi = -\frac{\KIforcei}{\Kforcei+1}\errorWrenchIntegrali +  \frac{\mass_i}{\Kforcei+1} \B\acci ,
	\end{equation}
	where the Jacobian here is just the scalar $\jacobianForce = -\frac{\KIforcei}{\Kforcei+1}$.
}
The respective \acp{LLE} $\lle_{\tau,i}$ can be computed for each first-order dynamics in \eqref{force_dynamics_matrix} as
\begin{equation}
	\label{wrench_lle}
	\lambda^*_{\tau,i} = \frac{\errorWrenchi}{\errorWrenchIntegrali} , \forall i \in \{1, \dots, 6\}.
\end{equation}
\rwSix{Notice that, since the systems in \eqref{impedance_dynamics_matrix} and \eqref{force_dynamics_matrix} are non-autonomous, the \ac{LLE} estimate will vary under the influence of external disturbances.}

\subsection{\ac{CBF}-based power flow control}
Consider a generic nonlinear system with dynamics
\begin{equation}
	\label{generic_system}
	\dot{\bfx} = \bff(\bfx) + \bfg(\bfx)\bfu ,
\end{equation}
where $\bfx \in \nR{n}$ and $\bfu \in \nR{m}$ are state and input respectively.
Let $h(\bfx,t)$ be a continuously differentiable scalar function of the system state (possibly time varying) and $\calC = \{ \bfx \in \nR{n} : h(\bfx)\geq 0  \}$ the zero superlevel set of $h$.
Then $h$ is a \ac{CBF} for the system in \eqref{generic_system} if there exists an extended class $\calK_\infty$ function $\gamma$ such that for the system \eqref{generic_system} it is
\begin{equation}
	\label{cbf_condition}
	\sup_{u \in \nR{m}} \left\{ \frac{\partial h}{\partial t} + \calL_{\bff} h(\bfx) + \calL_{\bfg} h(\bfx) \bfu + \gamma(h(\bfx))\right\}  \geq 0 ,
\end{equation}
\rwSix{where $\calL_{\bff} h(\bfx)$ and $\calL_{\bfg} h(\bfx)$ are the Lie derivatives of $h$ with respect to the vector fields $\bff$ and $\bfg$ respectively, as defined in \cite{isidori1995nonlinear}.}
Then, any Lipschitz continuous controller $\bfu(\bfx) \in \{ \bfu \in \nR{m} : \frac{\partial h}{\partial t} + \calL_{\bff} h(\bfx) + \calL_{\bfg} h(\bfx) \bfu + \gamma(h(\bfx)) \geq 0 \}$ renders the \textit{safe} set $\calC$ forward invariant \cite{Ames2019}.
\EC{
	The function $\gamma$ represents the distance of the system's state trajectory from the safe set boundary.
	The simplest choice for it is a proportional law $\gamma = k_\gamma h$, with gain $k_\gamma \in \nR{+}$.
	The higher the gain, the more the system dynamics are allowed to approach the set boundary.}
The \ac{CBF} condition in \eqref{cbf_condition} can be seen as a linear constraint on a control input $\bfu\refer$ inside a quadratic program (QP)
\begin{mini}|l|
	{\bfu}{\| \bfu - \bfu\refer \|^2}{}{}
	\label{qp_program}
	\addConstraint{\bfA(\bfx) \bfu}{\leq \bfb(\bfx,t)}{} ,
\end{mini}
where $\bfu\refer$ is a reference control input that comes, for instance, from a nominal controller and
\begin{subequations}
	\label{A_b_computation}
	\begin{IEEEeqnarray} {ll}
		\bfA(\bfx) &= -\calL_{\bfg} h(\bfx) ,\\
		\bfb(\bfx,t) &= \frac{\partial h}{\partial t} + \calL_{\bff} h(\bfx) + \gamma(h(\bfx)) .
	\end{IEEEeqnarray}
\end{subequations}

Let us define a constraint on the power injected by the pose or wrench tracking controllers into the system dynamics
\begin{equation}
	\label{power_limit}
	\errorVelStacked\transpose \wrenchCmd \leq \pLim(t),
\end{equation}
with $\pLim(t)$ a time-varying power limit\footnote{If the controller does not have a reference velocity, as in the case of wrench tracking, the velocity error $\errorVelStacked$ becomes just $\B\vel$.}.
This leads to the definition of a \ac{CBF} of the form $\powercbf = \pLim - \errorVelStacked\transpose \wrenchCmd$.
Notice that the control input $\wrenchCommand$ might be a function of the system's state variables, as in case of the pose tracking controller in \eqref{position_control}. However, this is not the case for the wrench tracking controller in \eqref{wrench_control} since it directly depends on the wrench measurements, coming from a \ac{FT} sensor.
Since \acp{CBF} need to be imposed on state variables, it is necessary to move the control action to the jerk level, considering the control input $\wrenchCommand$ as another system state.
\rwNine{This guarantees continuity of the control input even when switching between different controllers (pose and wrench tracking in this case).}
Following the work in \cite{Ames2021}, we augment the system dynamics in \eqref{omav_fb} as
\begin{equation}
	\label{omav_aug}
	\begin{cases}
		\totalInertia \B\acc = \wrenchCommandAugment + \wrenchExt \\
		\jerkCommand = \wrenchCommandDot + \Kjerk \( \wrenchCommand - \wrenchCommandAugment \)
	\end{cases},
\end{equation}
where $\Kjerk \in \nR{6 \times 6}$ is a diagonal and positive definite gain matrix that allows the control wrench $\wrenchCommandAugment$ of the augmented dynamics to track the nominal control wrench $\wrenchCommand$.
Notice that $\wrenchCommandDot$ can be easily computed analytically when $\wrenchCommand$ is only a function of the state, or numerically otherwise.
The QP computed on the augmented system is in the form
\begin{mini}|l|
	{\jerkCommand}{\| \jerkCommand - \jerkCommand\refer \|^2 + \slackWeight\norm{\slack}^2}{}{}
	\label{qp_jerk_program}
	\addConstraint{\bfA_p \jerkCommand + \slackPower}{\leq \bfb_p}{}
	\addConstraint{\bfA_{\tau} \jerkCommand + \slackInput}{\leq \bfb_{\tau}}{}
	\addConstraint{\eye{6}\jerkCommand}{\leq \jerkCommandLimit}{} ,
\end{mini}
where $\bfA_p \in \nR{6 \times 6}$ and $\bfb_p \in \nR{6}$ represent the power flow constraint in \eqref{power_limit}, while $\bfA_{\tau} \in \nR{6 \times 6}$ and $\bfb_{\tau} \in \nR{6}$ allow to enforce constraints on the input wrench $\wrenchCommandAugment$.
The last constraint on the jerk control input $\jerkCommand$ is imposed since aerial platforms can be particularly sensitive to high jerk commands.
Also, $\slack = \left[ \slackPower\transpose \vSpace \slackInput\transpose \right]\transpose \in \nR{12}$ represents a vector of slack variables
weighted by a very high coefficient $\slackWeight \sim 10^6$ in order to keep them as low as possible.
The addition of slack variables to the QP is necessary to guarantee the problem's feasibility in case of actuators saturation.
\EC{
	To impose an input wrench limit $\norm{\tau_{a,i}}\leq\bar{\tau}_{a,i} \forall i \in \{1,\dots,6\}$, we choose as \acp{CBF} $\inputcbf{}_{,i} = \bar{\tau}_{a,i}^2 - \tau_{a,i}^2 $.
	The constraints in \eqref{qp_jerk_program} can be then computed from \eqref{A_b_computation} considering the desired \acp{CBF} for power $\powercbf{}_{,i}$ and control wrench $\inputcbf{}_{,i}$:
}
\begin{subequations}
	\label{qp_quantities}
	\begin{IEEEeqnarray} {ll}
		\bfA_p & = \text{diag}\{\errorVelStackedOne , \dots , \errorVelStackedSix\}\\
		b_{p,i} & = \powercbfGain{}_{,i} \powercbfi + \dot{\pLim}_i - \frac{1}{\mass_i}\wrenchCommandAugmenti \(\wrenchCommandAugmenti + \wrenchExti\)\\
		\bfA_{\tau} & = 2\; \text{diag}\{\wrenchCommandAugmenti , \dots , \wrenchCommandAugmenti\}\\
		b_{\tau,i} & = \inputcbfGain{}_{,i} \inputcbfi .
	\end{IEEEeqnarray}
\end{subequations}
Since $\bfA_p$ and $\bfA_{\tau}$ are diagonal, power flow and control input constraints can be independently imposed for each \ac{DoF}.

\subsection{Power adaptation law}
We want to dissipate power when the closed loop dynamics start to diverge ($\lle > 0$).
At the same time, we allow for power generation when the dynamics are converging ($\lle < 0$).
One possible adaptation law could just be proportional $\pLim = - \Klamb \lle$.
Since the imposed power dissipation must always be feasible \EC{(it is not possible to dissipate if the system has no energy left)}, we choose to adapt the power flow limit $\pLim(\lle,\errorVeli)$ as
\begin{equation}
	\label{power_law}
	\begin{cases}
		\pLim = - \Klamb \lle              & \text{if}\; \lle < 0 \\
		\pLim = - \Klamb \lle \errorVeli^2 & \text{if}\; \lle > 0
	\end{cases}
	.
\end{equation}
This way, the gain $\Klamb$ represents a damping coefficient when a power dissipation is required.

\section{Safe sets analysis}
\label{Par:safe_sets}
The presented safety layer constrains the system's evolution into the safe sets $\PosSafeSet$ and $\ForceSafeSet$ for the position and wrench tracking dynamics, respectively.
Importantly, we show which limitations these sets impose and that the control objective is always reachable even when these limits are enforced.
\subsection{Pose tracking set}
\begin{figure}[t]
	\centering
	\begin{subfigure}{0.352\columnwidth}
		\includegraphics[width=0.98\columnwidth]{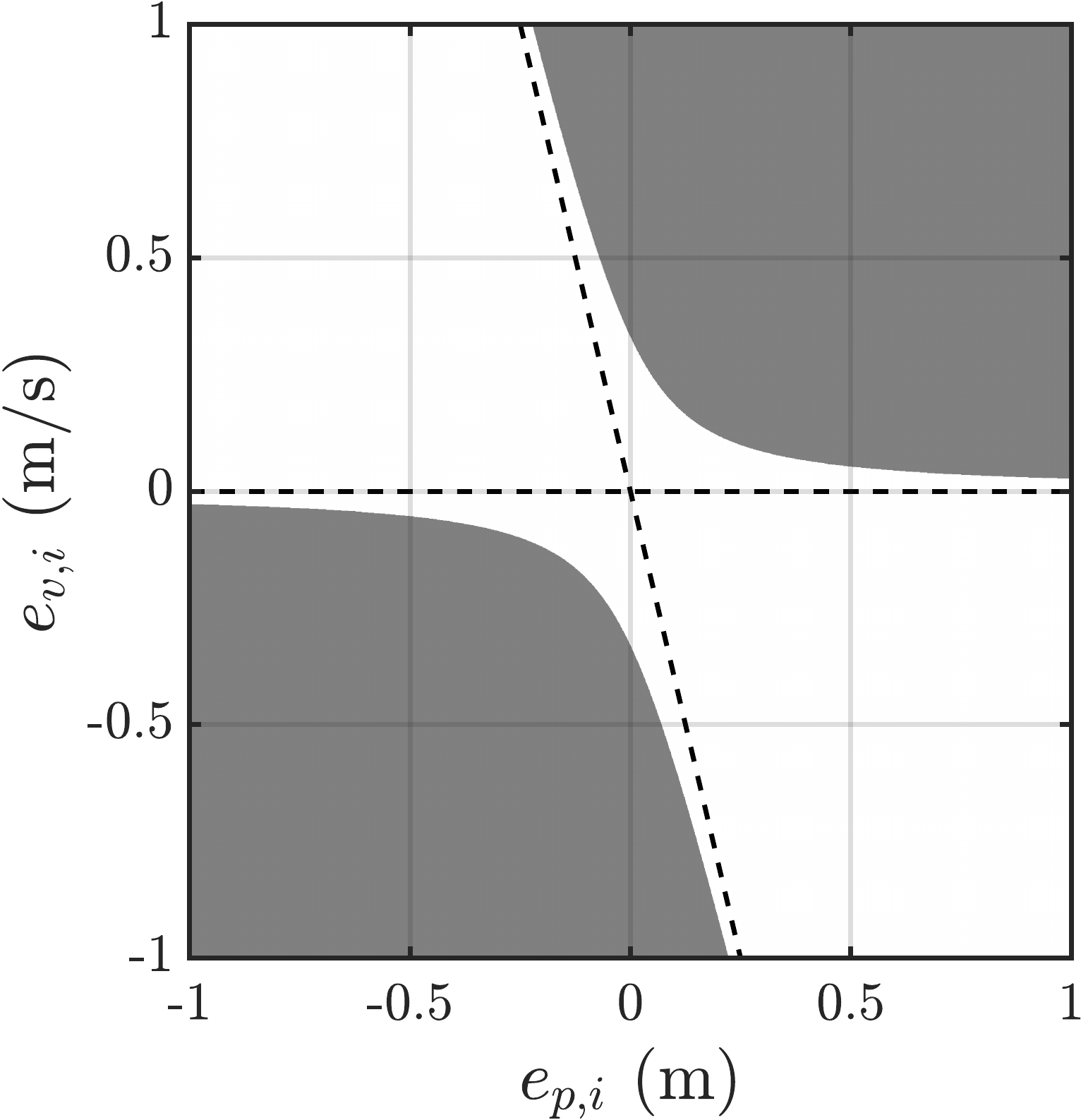}
	\end{subfigure}%
	\begin{subfigure}{0.31\columnwidth}
		\includegraphics[width=0.98\columnwidth]{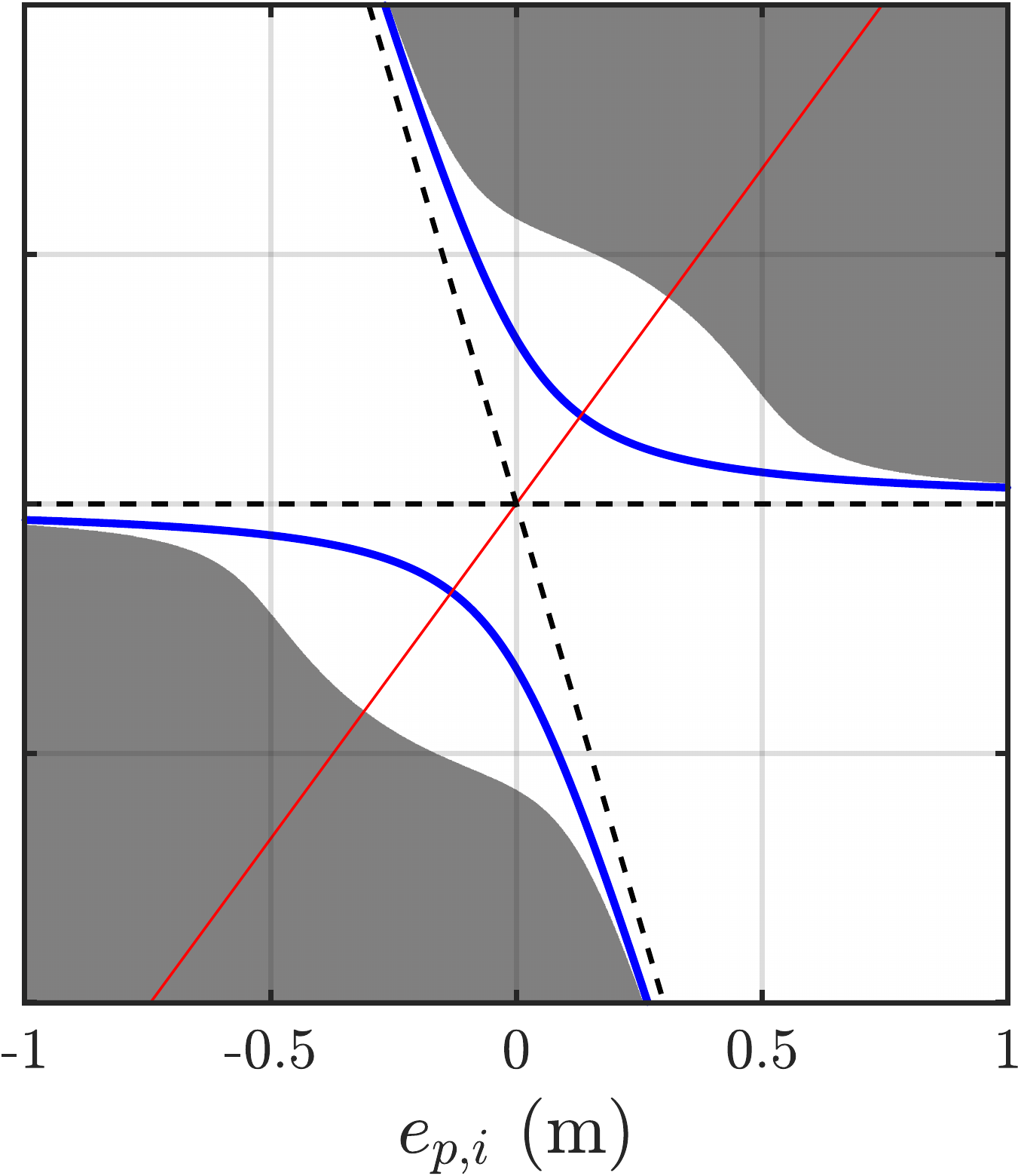}
	\end{subfigure}%
	\begin{subfigure}{0.31\columnwidth}
		\includegraphics[width=0.98\columnwidth]{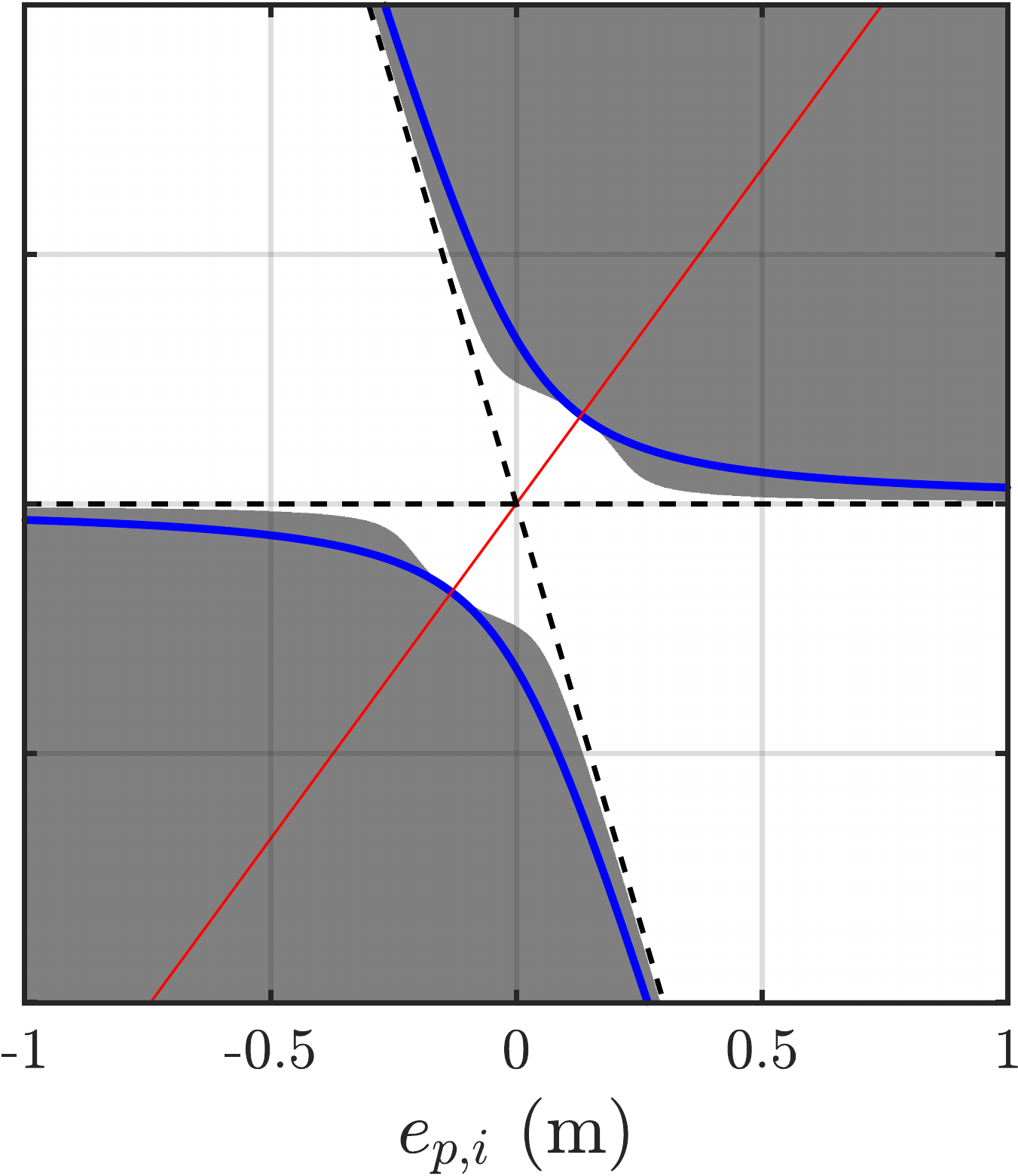}
	\end{subfigure}
	\caption{On the left, forward invariant safe set (white) for the system's position dynamics. In the center, the set without acceleration measurements exceeds the original set. On the right, the set is scaled to fit inside the original one. The dashed lines represent the asymptotes of the set. The red line represents the scaling direction. Sets obtained obtained considering $\mass_i = \SI{4.58}{\kilogram}$, $\Kdampingi=\SI{5}{\newton\per\metre\per\second}$, $\Kspringi=\SI{20}{\newton\per\metre}$ and $\Klamb=\SI{1}{\watt\second}$.}
	\label{fig:pose_accel_set}
\end{figure}
The safe set for the decoupled pose tracking controller dynamics is given by $\PosSafeSet = \{ (\errorPosStackedi, \errorVelStackedi) \in \nR{} \times \nR{} : \errorVelStackedi \wrenchPosContri \leq \pLim(\lle,\errorVeli) \}$, where $\wrenchPosContri$ represents the corresponding  element of the pose control vector in \eqref{position_control}.
By considering the position dynamics only, the obtained safe set is shown on the left in Fig.~\ref{fig:pose_accel_set}.
\EC{From the definition of $\PosSafeSet$, considering \eqref{position_control} and \eqref{power_law}, one can find the boundaries of the pose safe set to be an hyperbola in the state space}, with asymptotes $\errorVelStackedi = 0$ and $\errorVelStackedi = - \frac{\Kspringi}{\Kdampingi}\errorPosStackedi$.
Fig.~\ref{fig:pose_accel_set} shows how the power flow adaptation law in \eqref{power_law} represents a safe set that contains the equilibrium at the origin $(\errorPosStackedi, \errorVelStackedi) = (0,0)$.
Moreover, the safe set cuts out of the feasible space most of the bottom-left and top-right quadrants, in which the position and velocity errors are aligned.
This is the most dangerous situation according to the \ac{LLE} metric, in which the position error norm is increasing.

In case the \ac{LLE} is computed using \eqref{impedance_lle} but the acceleration error is approximated to be $\errorAccStackedi=0$, the real value of the acceleration results in a mismatch on the $\lle$ estimation.
This causes an enlargement of the safe set around the origin compared to the original one, as shown in Fig.~\ref{fig:pose_accel_set}.
\EC{This especially expands in the dangerous regions in which the position and velocity errors are aligned.
	In these regions, from Fig.~\ref{fig:pose_accel_set} we see an increase in the maximum velocity of around double the original value (from $\sim\SI{0.25}{\metre\per\second}$ to $\sim\SI{0.5}{\metre\per\second}$).
	This translates to a possible increase of the system kinetic energy and thus a reduction of the controller safety.
}
Since the enlarged set still resembles an hyperbola with the same asymptotes of the nominal one, it might be scaled to fit inside the latter.
The scaling can be performed by imposing stricter limits on the power flow as $\pLim_s = \KlambStrict \pLim$, where $\pLim_s$ is the new imposed power limit and $\KlambStrict < 1$.
The direction in which the biggest enlargement happens is the bisector of the two asymptotes, which has equation $\errorVelStackedi = \frac{1}{\Kspringi} \left(\Kdampingi - \sqrt{\Kspringi^2 + \Kdampingi^2}\right)\errorPosStackedi$.
By enforcing the two sets to match along this direction, one finds $\KlambStrict = -\frac{\lle}{c_1 - \lle}$, with
\begin{equation}
	c_1 = \frac{1}{2\mass_i}\left( \Kdampingi + \sqrt{(\Kdampingi^2 + \Kspringi^2)} \right).
\end{equation}

\subsection{Wrench tracking set}
Since we are interested on the dynamics of the platform under the wrench tracking control action, we define the wrench control safe set in terms of the platform speed and wrench as $\ForceSafeSet = \{ (\wrenchForceContri, \B\veli) \in \nR{} \times \nR{} : \B\veli \wrenchForceContri \leq \pLim(\lle,\B\veli) \}$.
\begin{figure}[t]
	\centering
	\begin{subfigure}{0.5\columnwidth}
		\includegraphics[width=0.98\columnwidth]{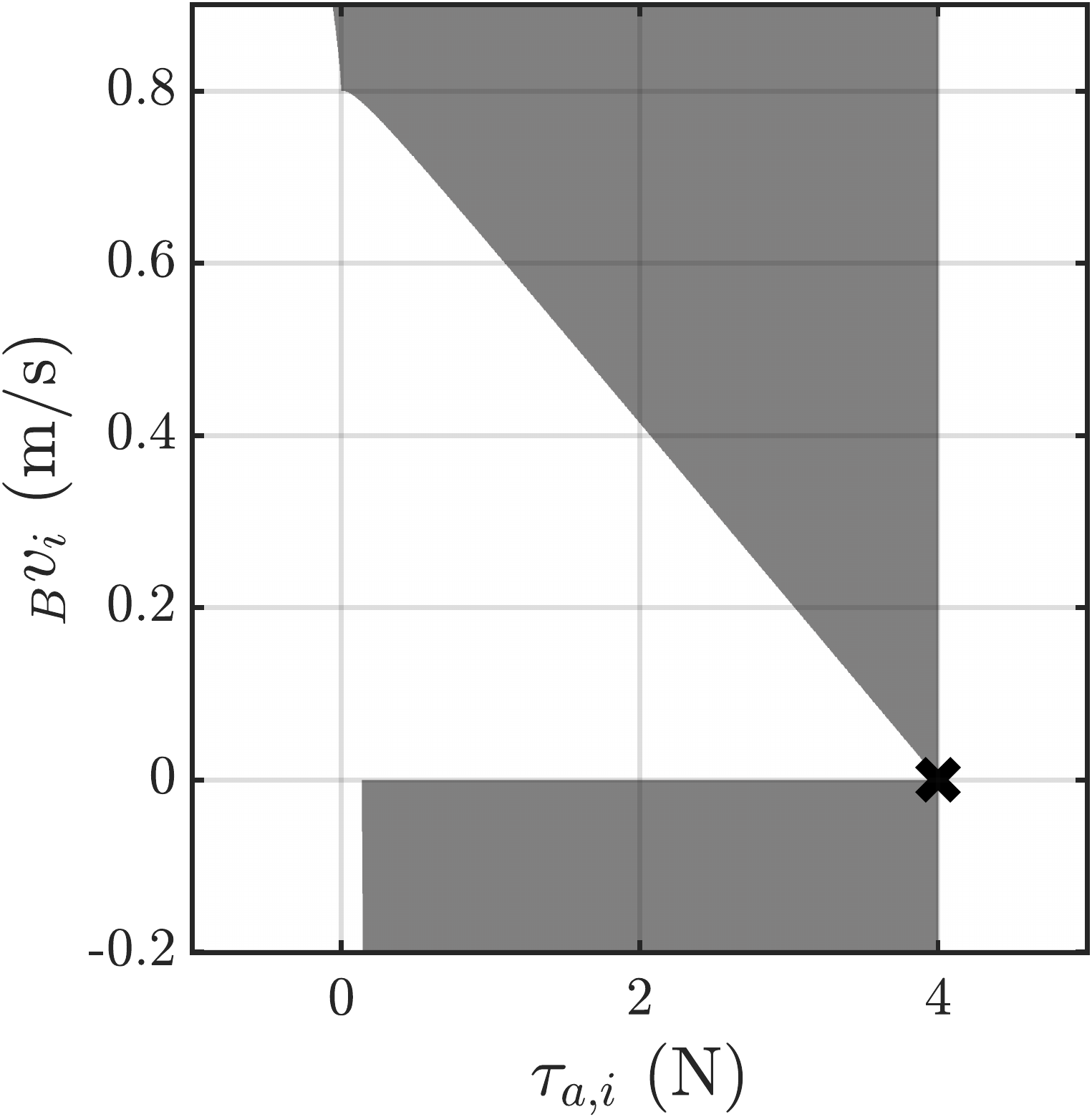}
	\end{subfigure}%
	\begin{subfigure}{0.5\columnwidth}
		\includegraphics[width=0.98\columnwidth]{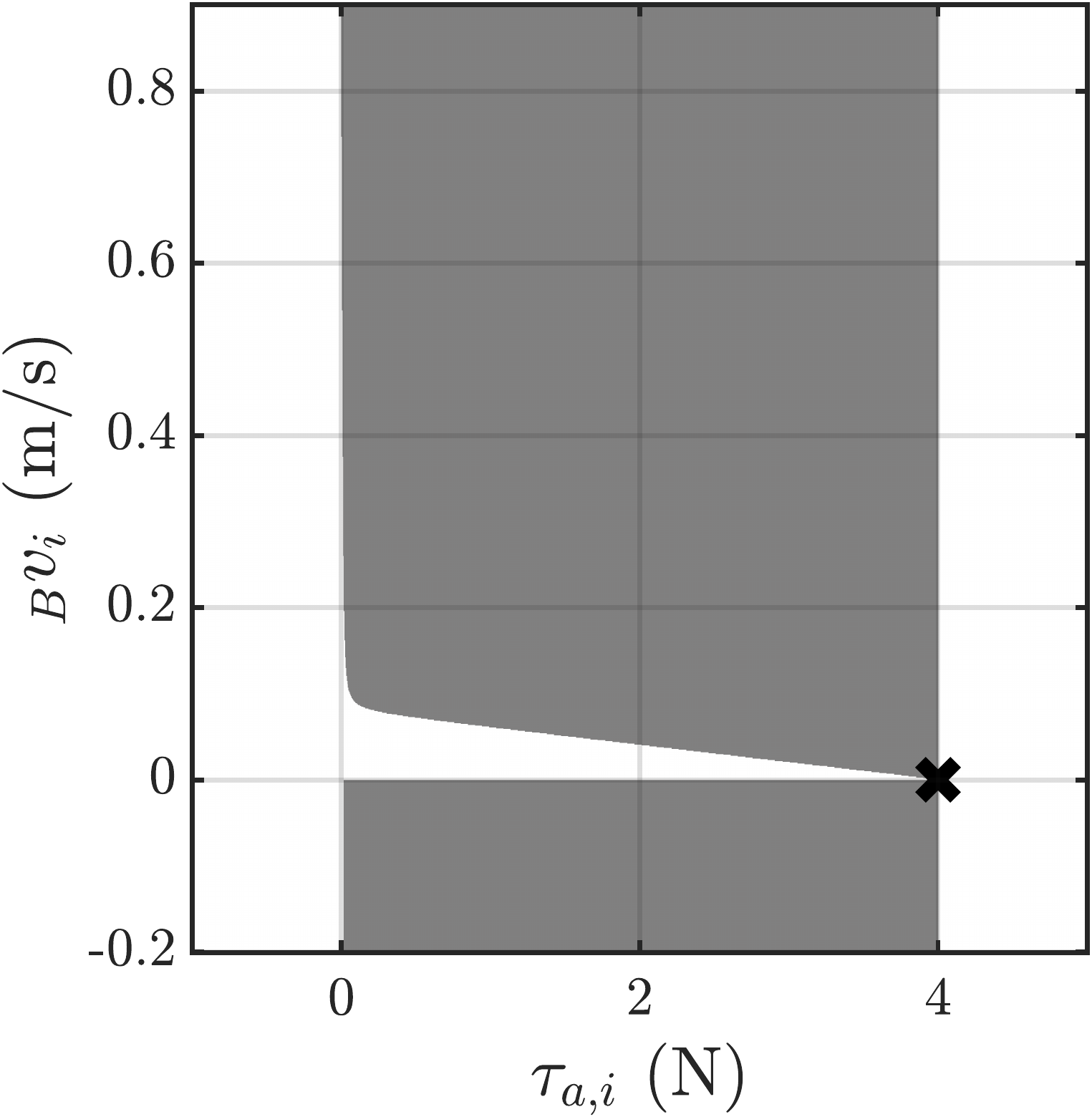}
	\end{subfigure}
	\caption{Safe sets considering a force setpoint $\wrenchExt\refer = 4 N$ (indicated with a black cross) with a surface with damping coefficient $\Kdwalli=\SI{5}{\newton\per\meter\per\second}$ and stiffness coefficient $\Kpwalli=\SI{30}{\newton\per\meter}$ (left) or $\Kpwalli=\SI{300}{\newton\per\meter}$ (right). The power coefficient is $\Klamb=\SI{1}{\watt\second}$.}
	\label{fig:force_accel_set}
\end{figure}
First, consider the case of activation of the \ac{WTC} \EC{when there is no interaction surface}, or the case in which the interaction surface moves away making the wrench tracking not possible anymore.
Without the presented safety layer, the wrench tracking errors would then increase causing the platform to accelerate in the desired force direction and diverge.
With the proposed power limitation, as soon as the force dynamics start to diverge, the correspondent \ac{LLE} becomes positive, enforcing a dissipative behavior and eventually stopping the platform.
When this is the case, no wrench reference is possible to be achieved and the safe set collapses to the axis $\B\veli = 0$.
On the other hand, when there is a surface on which the platform can push, the safe set assumes the shape in Fig.~\ref{fig:force_accel_set}.
\EC{To obtain this figure it is necessary to specify a model for the environment such to define an interaction force feedback $\wrenchExt$ on the vehicle.
	In this case we suppose to interact with a fixed environment, whose restitution force we model as $\wrenchExti = -\Kdwalli \B\veli - \Kpwalli \errorPosWallStackedi $, with $\errorPosWallStackedi$ elements of a position error vector $\errorPosWallStacked \in \nR{6}$ between the vehicle and the surface.
}
The set in Fig.~\ref{fig:force_accel_set} depends on the damping and stiffness characteristics of the environment but the peculiarity is that the force setpoint is always part of the set and thus reachable.
Moreover, the system's speed is limited more and more as we approach the desired force.
By changing the surface parameters, specifically when increasing the ratio between stiffness and damping, the gap of allowed velocity narrows, but the desired force remains feasible.
\EC{This might be explained by the higher restitution force of the surface.
	With higher forces applied on the platform, the safety layer would impose stricter limits on the system's velocity in order to keep the power flow limited.}
\EC{Notice that the specific surface properties are here used just to give an hint of what effect these power limitations have while in interaction.
	They are not necessary to implement the proposed safety framework.}
\section{Experimental validation}
This section presents an experimental validation of the presented safety layer.
The proposed approach will be tested both in free flight  and interaction experiments.

\subsection{Free flight disturbances}
\begin{figure}[!tb]
	\centering
	\begin{subfigure}{0.5\columnwidth}
		\includegraphics[width=0.98\columnwidth]{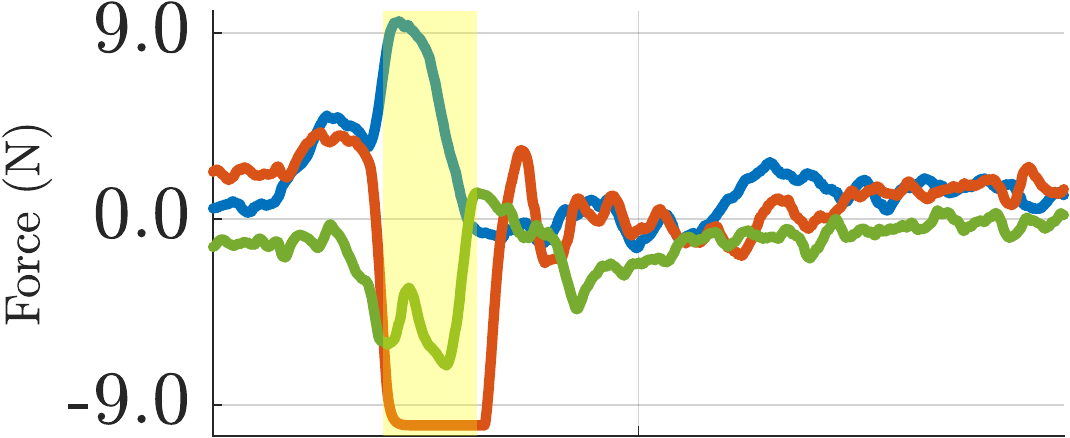}
	\end{subfigure}%
	\begin{subfigure}{0.5\columnwidth}
		\includegraphics[width=0.98\columnwidth]{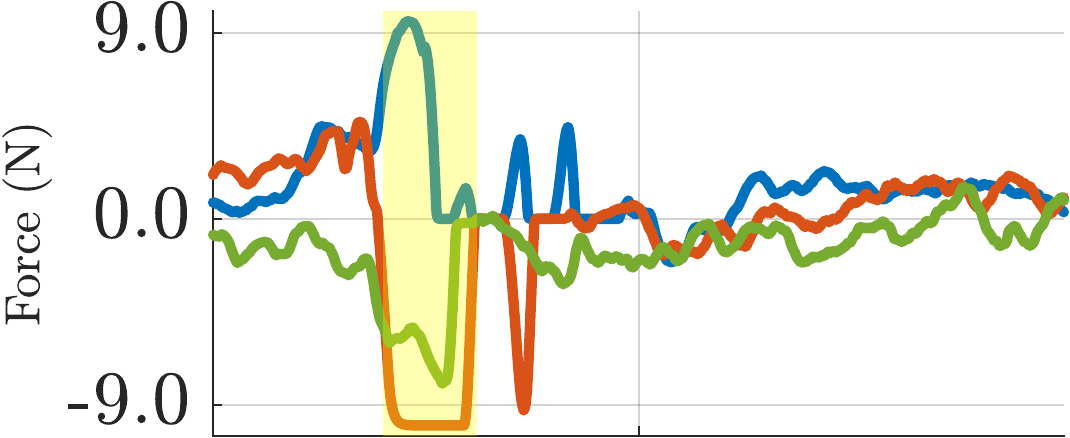}
	\end{subfigure}
	\begin{subfigure}{0.5\columnwidth}
		\includegraphics[width=0.98\columnwidth]{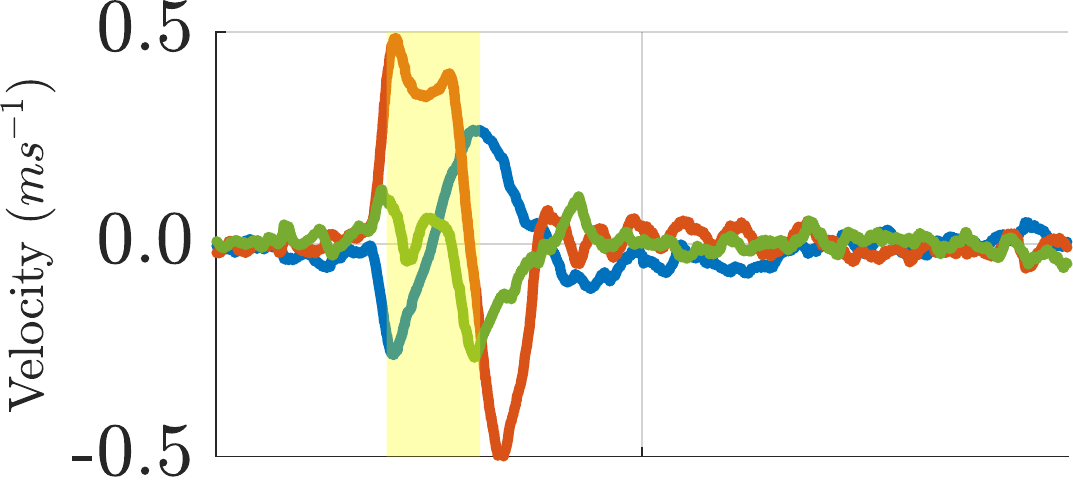}
	\end{subfigure}%
	\begin{subfigure}{0.5\columnwidth}
		\includegraphics[width=0.98\columnwidth]{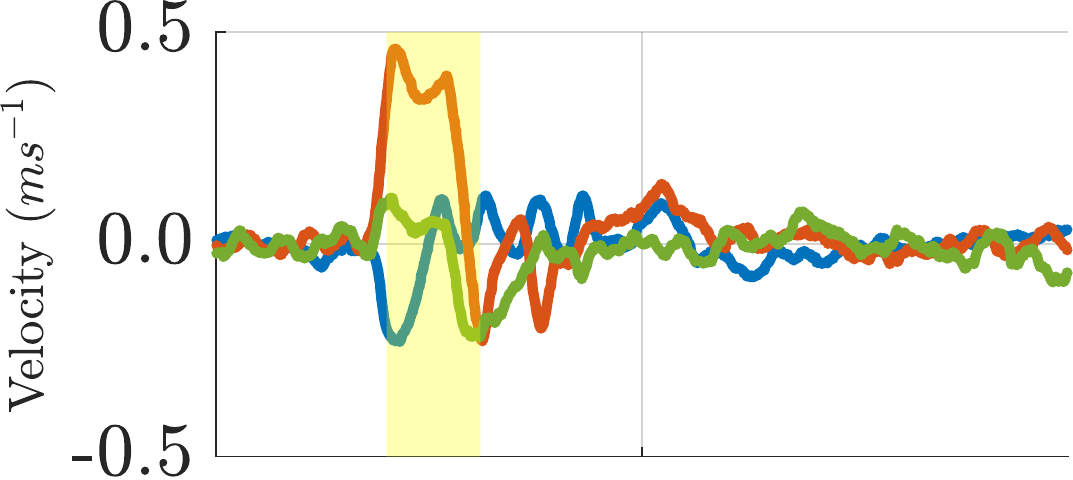}
	\end{subfigure}
	\begin{subfigure}{0.5\columnwidth}
		\includegraphics[width=0.98\columnwidth]{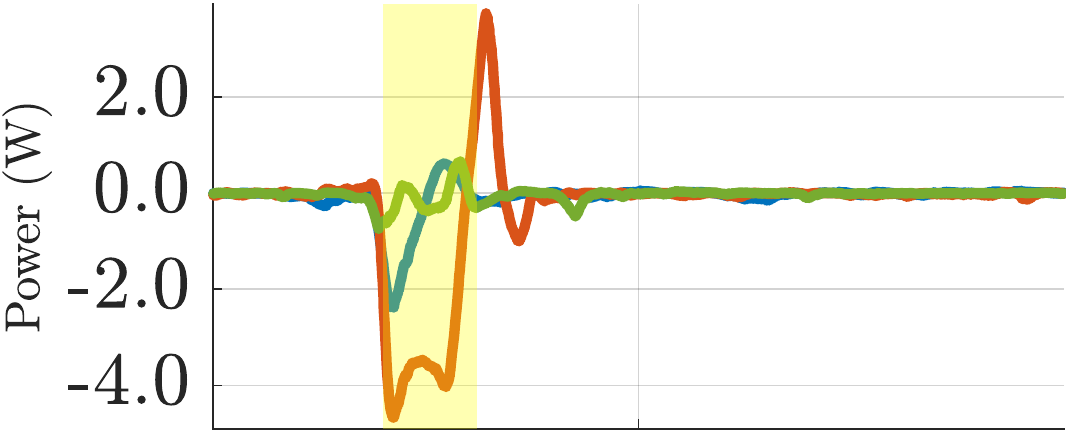}
	\end{subfigure}%
	\begin{subfigure}{0.5\columnwidth}
		\includegraphics[width=0.98\columnwidth]{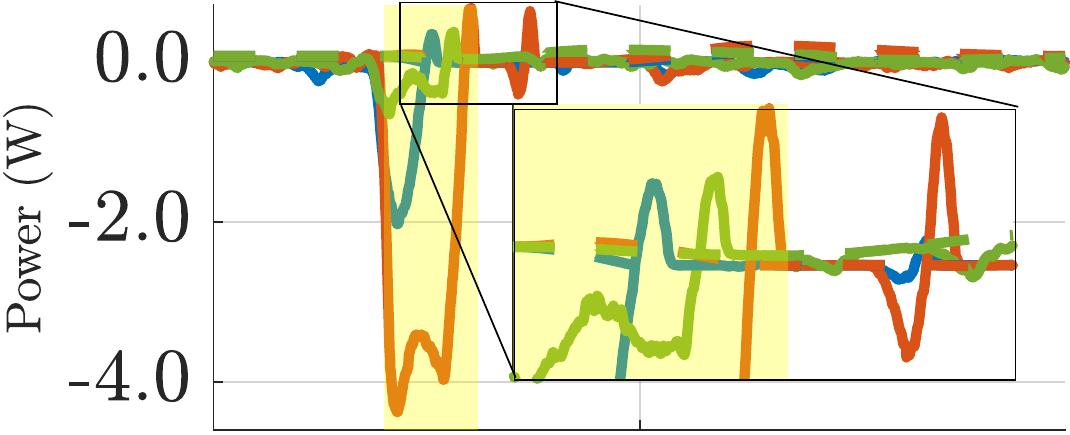}
	\end{subfigure}
	\begin{subfigure}{0.5\columnwidth}\hfill
		\includegraphics[width=0.98\columnwidth]{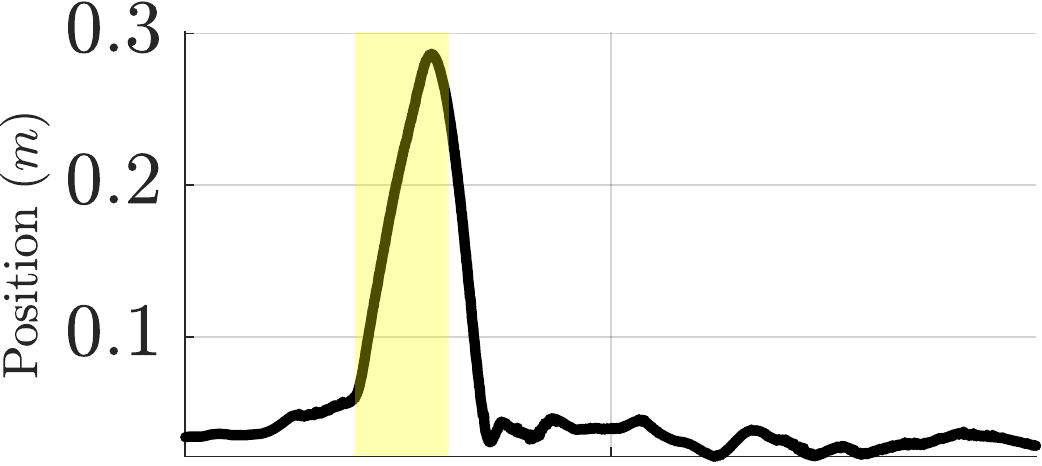}
	\end{subfigure}%
	\begin{subfigure}{0.5\columnwidth}\hfill
		\includegraphics[width=0.98\columnwidth]{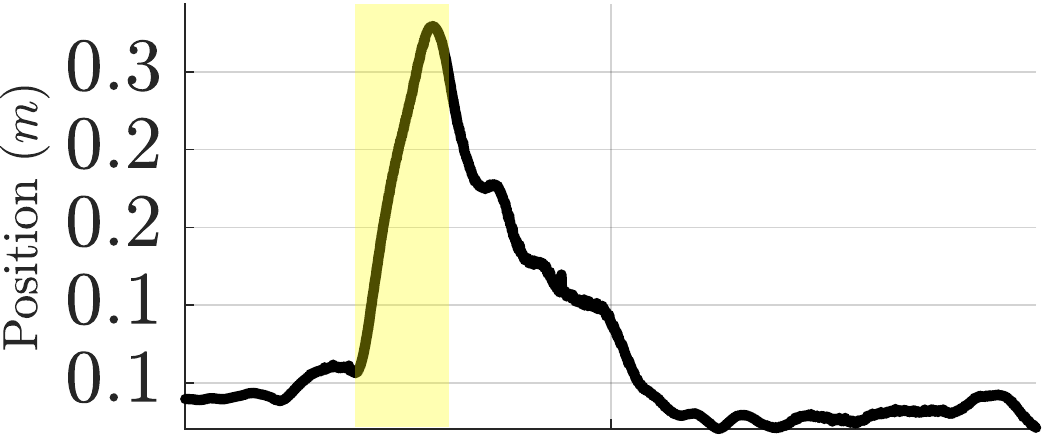}
	\end{subfigure}
	\begin{subfigure}{0.5\columnwidth}
		\includegraphics[width=0.98\columnwidth]{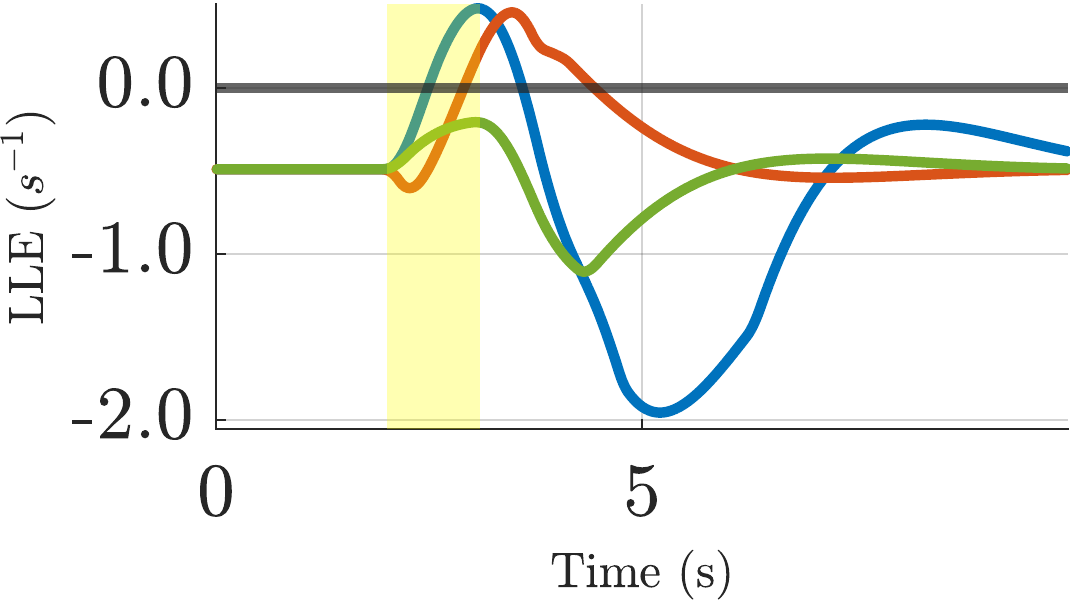}
	\end{subfigure}%
	\begin{subfigure}{0.5\columnwidth}
		\includegraphics[width=0.98\columnwidth]{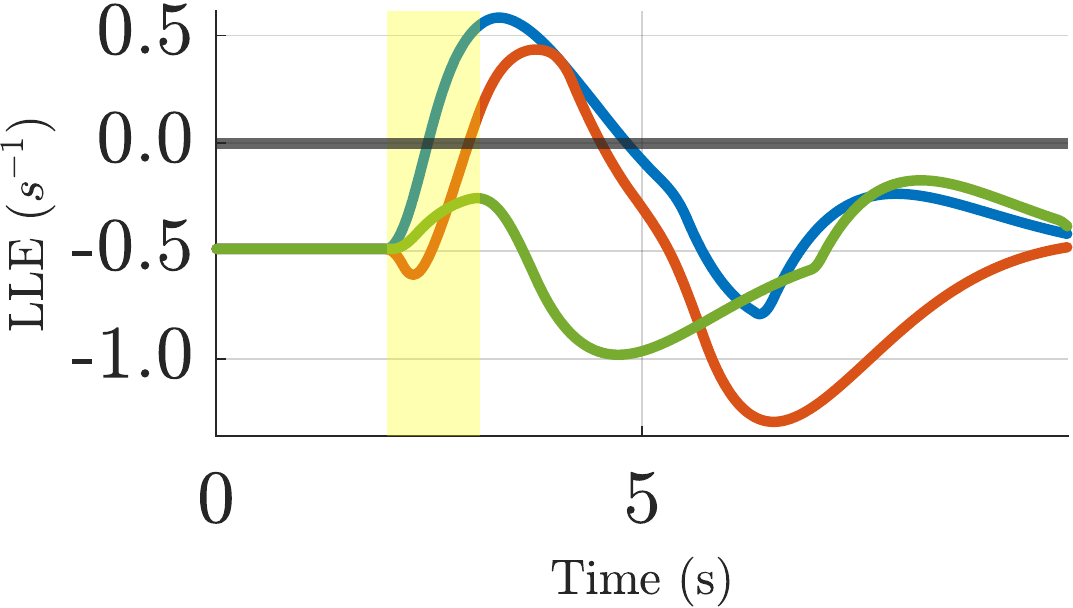}
	\end{subfigure}
	\caption{\rwThree{Pose tracking experiment without (left column) and with (right column) safety framework. From top to bottom, the rows show force control inputs, trajectory velocity errors, system's power flow, trajectory position errors norm, \ac{LLE} estimates, respectively. The linear quantities are displayed along the $x$ (blue), $y$ (orange) and $z$ (green) body axes. The disturbance time window is highlighted in yellow.}}
	\label{fig:pose_tracking_policy_comparison}
\end{figure}
\begin{figure}[!tb]
	\centering
	\begin{subfigure}{0.5\columnwidth}
		\includegraphics[width=0.98\columnwidth]{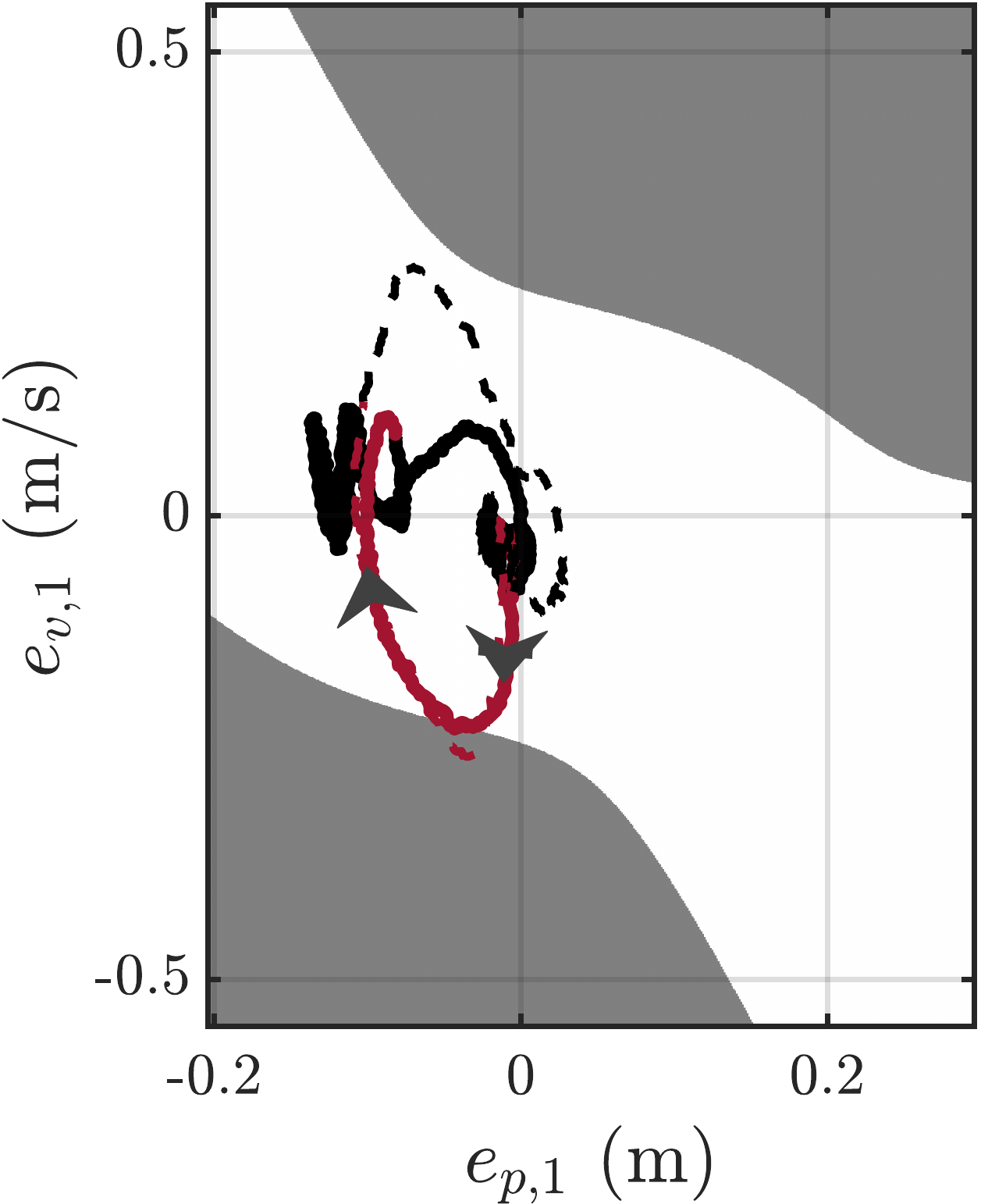}
		\caption{$x$ body axis.}
		\label{fig:safe_set_x}
	\end{subfigure}%
	\begin{subfigure}{0.5\columnwidth}
		\includegraphics[width=0.98\columnwidth]{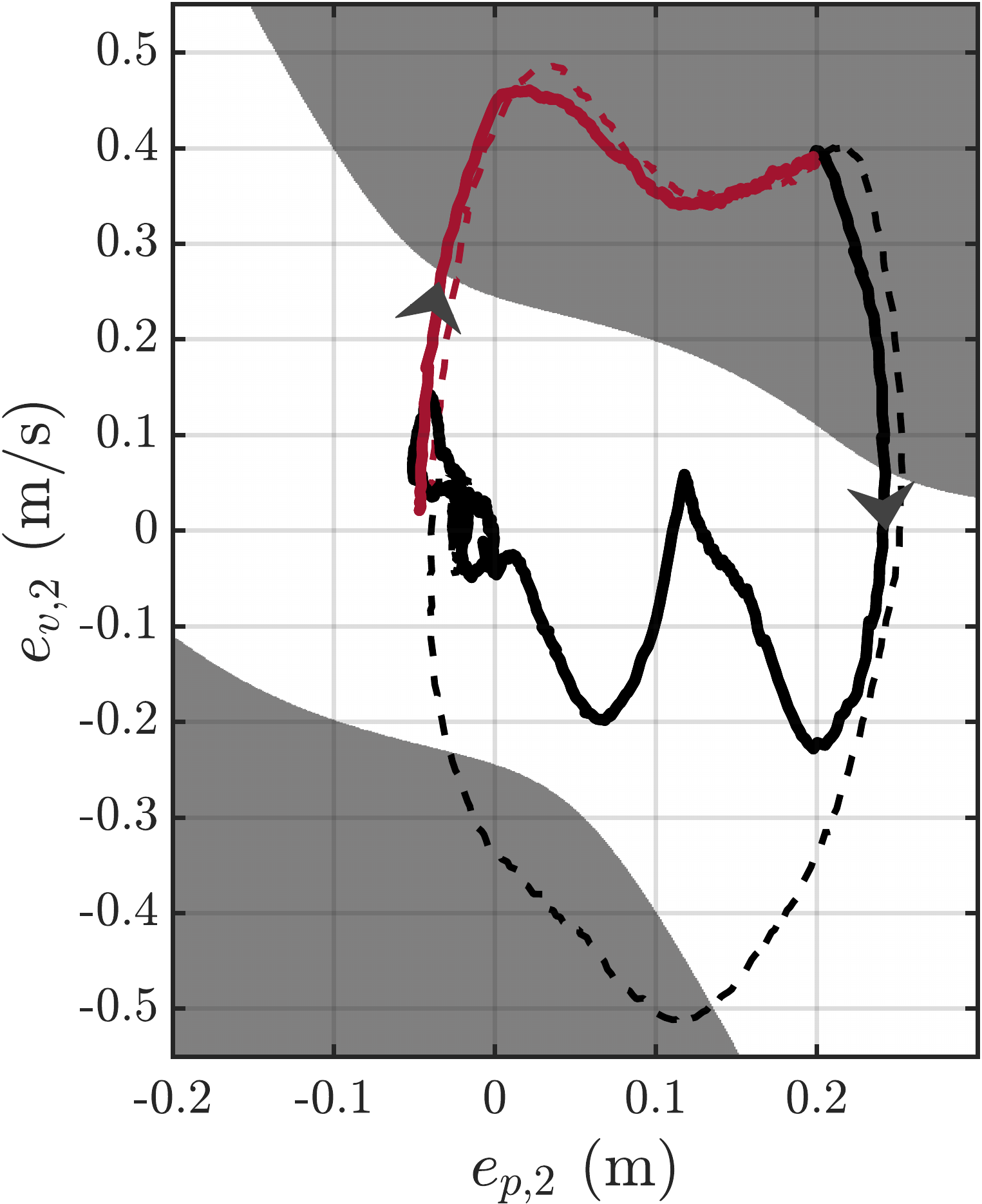}
		\caption{$y$ body axis.}
		\label{fig:safe_set_y}
	\end{subfigure}
	\caption{\rwThree{Safe set and position error trajectories with safety layer enabled (solid line) and disabled (dashed line). Time flows in the arrow's direction. The line portion in red represents the trajectory's part in which the disturbance was applied.}}
	\label{fig:exp_safety_set}
\end{figure}
\begin{figure*}[!htb]
	\centering
	\begin{subfigure}{0.32\textwidth}
		\includegraphics[width=0.98\columnwidth]{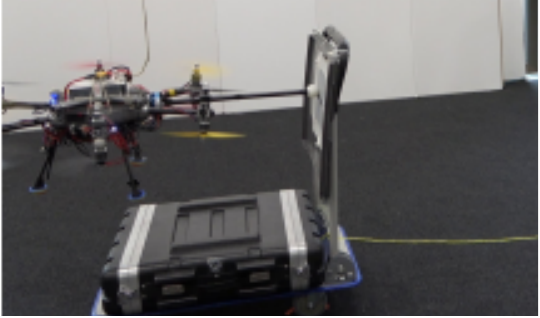}
		\caption{The aerial vehicle regulates a pushing force\\ on the cart.}
	\end{subfigure}%
	\begin{subfigure}{0.32\textwidth}
		\includegraphics[width=0.98\columnwidth]{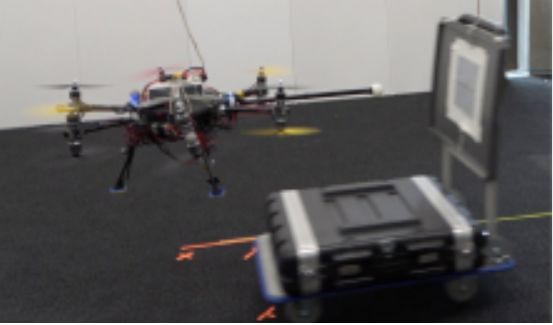}
		\caption{The cart is pulled away and the platform\\ loses contact.}
	\end{subfigure}%
	\begin{subfigure}{0.32\textwidth}
		\includegraphics[width=0.98\columnwidth]{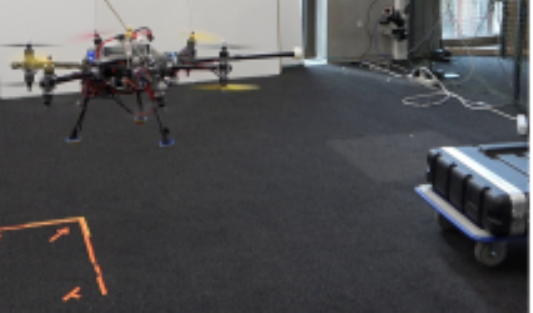}
		\caption{The force dynamics divergence is detected\\ and the platform stops without diverging.}
	\end{subfigure}
	\caption{Interaction experiment between the aerial platform and a moving cart.}
	\label{fig:interaction_exp_teaser}
\end{figure*}
In this experiment, the aerial platform is commanded to follow a \textit{figure eight} trajectory in free flight while subject to sudden virtual external disturbances.
\rwNine{As we only require position tracking, the selection matrix in \eqref{selected_control} is chosen as $\selectionMat=\mat{O}_{6\times6}$.}
\rwSix{We choose to apply virtual software disturbances since these are more controlled and repeatable for evaluation purposes. In reality, this could correspond to a sudden gust of wind or an impact.}
The disturbance is a \SI{14}{\newton} force in the world frame $\frameW$ $y$ axis.
This is applied on the system's \ac{COM} with the duration of \SI{1}{s}.
The linear control inputs are limited to $\norm{\wrenchCommandAugmenti} \leq \SI{10}{\newton} \vSpace \forall i \in \{1,  \dots, 3\}$ with the control input \ac{CBF}.
This was chosen in order to test the robustness of the proposed safety layer under actuator saturations.
The effects of these disturbances while the safety framework is enabled or not are visible in Fig.~\ref{fig:pose_tracking_policy_comparison}.
During the application of the disturbance, the drone's control inputs in the $x$-$y$ body plane saturate and the system is moved away from the planned trajectory.
When the disturbance suddenly stops, if the safety policy is not enabled, the pose tracking controller tries to bring the system back as fast as possible, causing it to overshoot and reach a velocity of \SI{0.5}{\metre\per\second} and \SI{0.25}{\metre\per\second} in the $y$ and $x$ direction, respectively.
Also the power generation reaches a value of \SI{4}{\watt} during the overshoot.
On the other hand, when the safety policy is enabled, we see the \acp{LLE} of the body linear dynamics increasing and becoming positive for the divergent directions.
The positive value of the \acp{LLE} enforces the system to limit and dissipate power, greatly reducing the overshoot once it goes back to the followed trajectory.
The maximum overshoots in the velocity errors $x$ and $y$ are now more than halved with respect to the previous case without safety policy.
Also the power flow generation is greatly reduced with maximum spikes of \SI{0.5}{\watt}.
Notice that the two positive power spikes in the $y$ direction around \SI{4}{\second} are actually given by the system breaking the power boundaries imposed by the \ac{CBF}.
This happens because the system saturates the relative control input.
\rwThree{Nonetheless, the power and the velocity tracking errors are greatly reduced, while the position tracking errors take around \SI{2}{\second} more to converge because of the damping imposed by the safety layer.
}
Also, when not saturating, the power flow goes back into the imposed boundaries, showing how the framework is robust to short power flow violations.
In the end, we also show in Fig.\ref{fig:exp_safety_set} the state trajectory in the safety set depicted in sec.~\ref{Par:safe_sets}.
The initial violation of the safe set is given by the disturbance and the corresponding saturation of the control input.

\subsection{Wrench tracking control} \label{wrench_tracking_subsection}
\begin{figure}[!tb]
	\captionsetup[subfigure]{width=0.9\linewidth}
	\begin{subfigure}{0.5\columnwidth}
		\includegraphics[width=0.98\columnwidth]{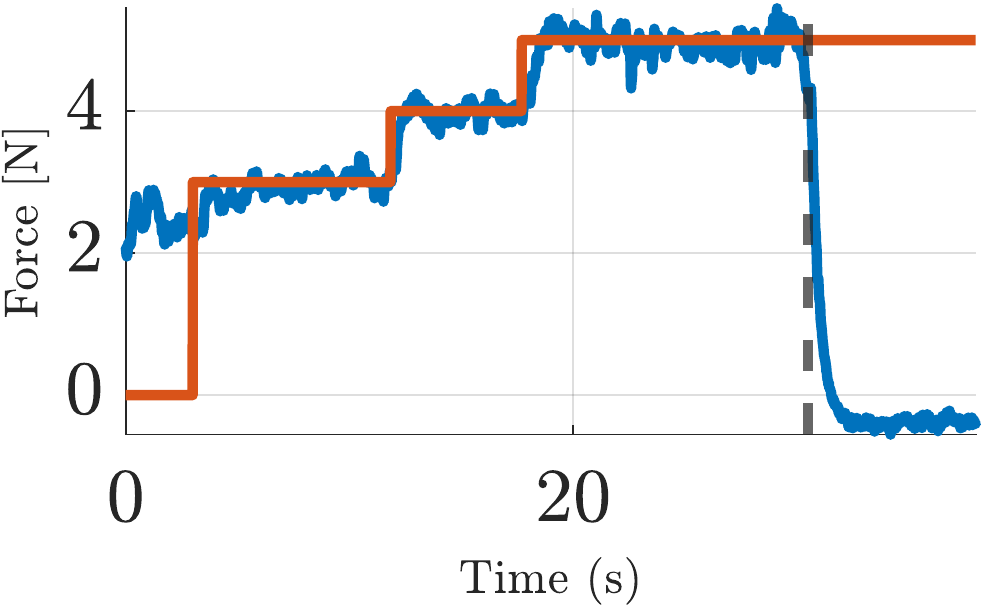}
		\caption{Measured force (blue) and force reference (orange).}
		\label{fig:wrench_tracking_policy_force}
	\end{subfigure}%
	\begin{subfigure}{0.5\columnwidth}
		\includegraphics[width=0.98\columnwidth]{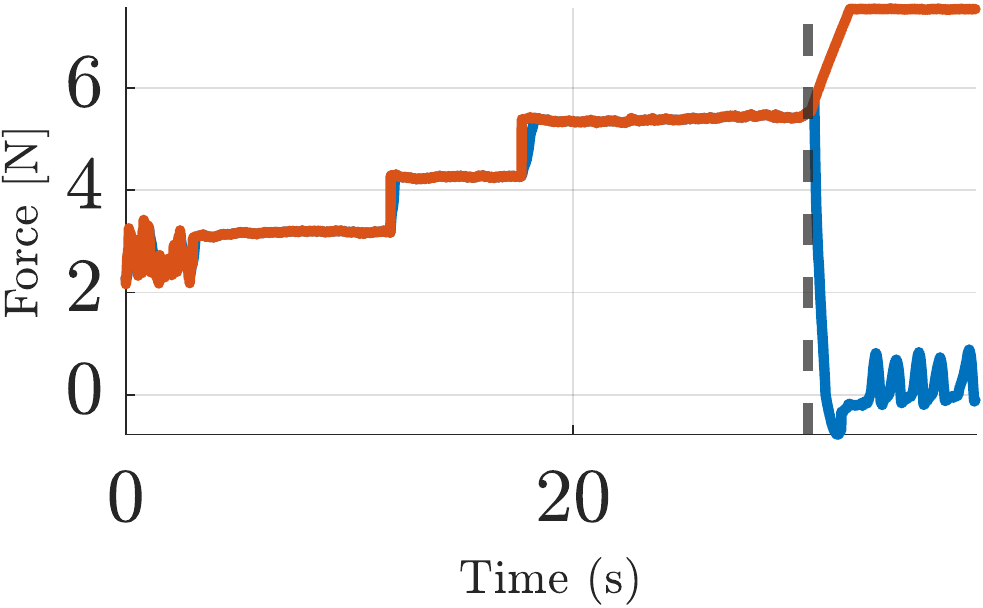}
		\caption{Desired force command $\wrenchCommand$ (orange) and safety filtered command $\wrenchCommandAugment$ (blue).}
		\label{fig:wrench_tracking_policy_command}
	\end{subfigure}

	\begin{subfigure}{0.5\columnwidth}
		\includegraphics[width=0.98\columnwidth]{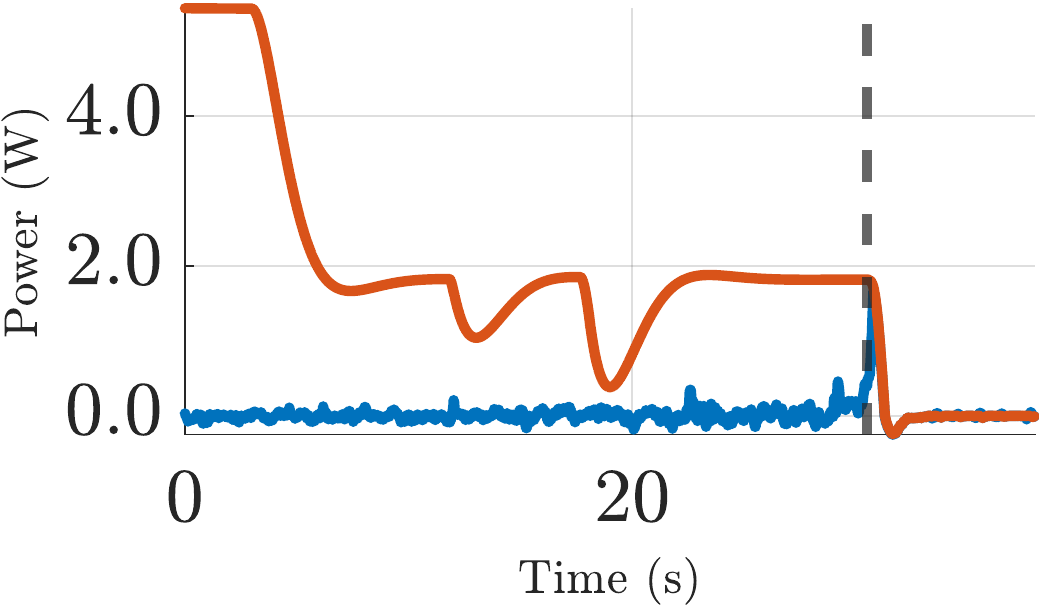}
		\caption{Power flow value (blue) and imposed power flow limit (orange).}
		\label{fig:wrench_tracking_policy_power}
	\end{subfigure}%
	\begin{subfigure}{0.5\columnwidth}
		\includegraphics[width=0.98\columnwidth]{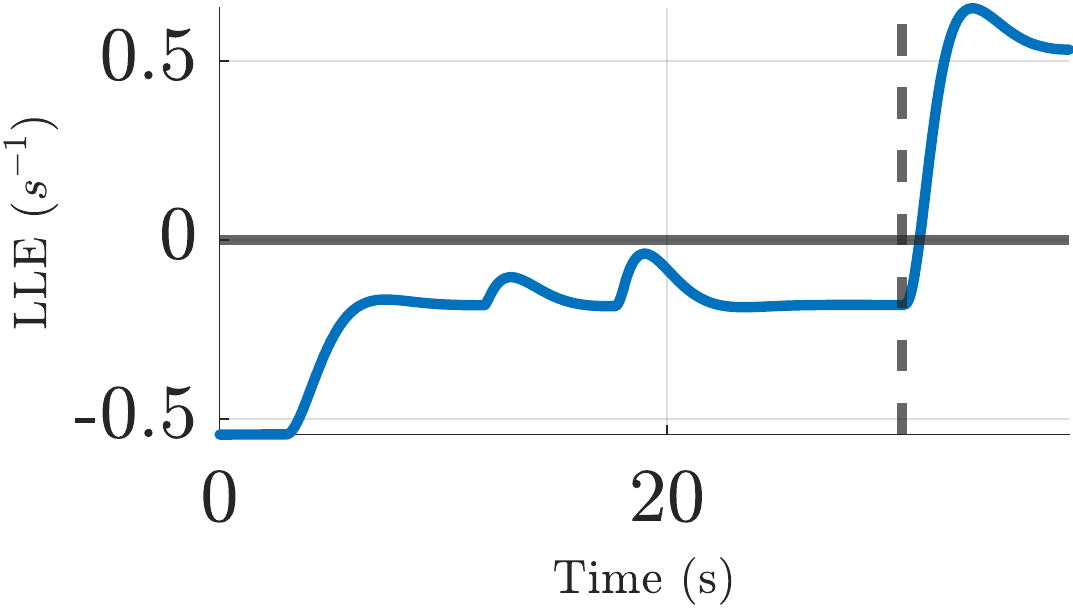}
		\caption{LLE estimation.}
		\label{fig:wrench_tracking_policy_lle}
	\end{subfigure}

	\begin{subfigure}{0.5\columnwidth}
		\includegraphics[width=0.98\columnwidth]{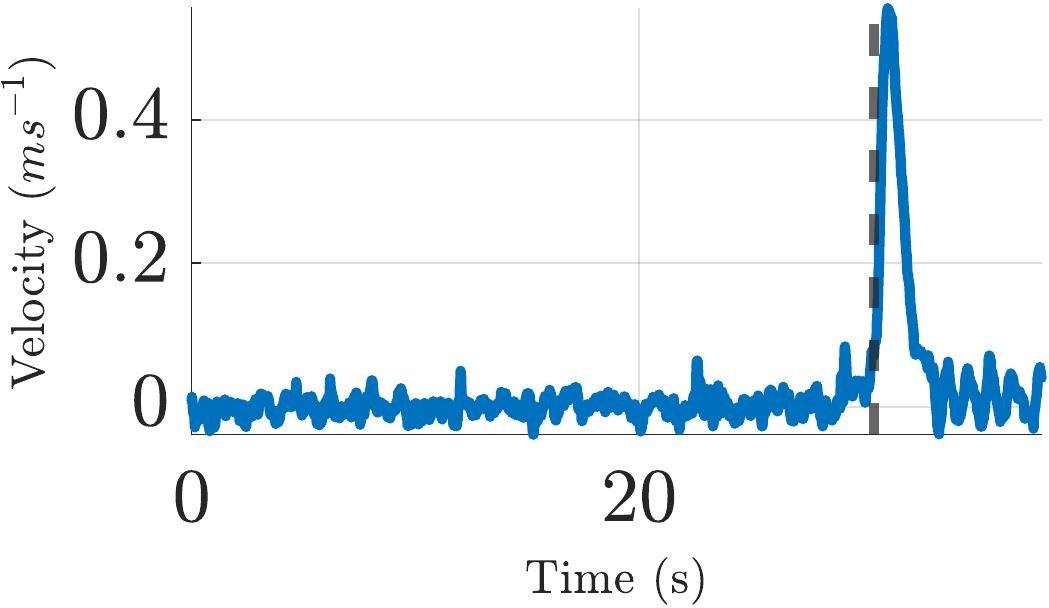}
		\caption{Linear velocity.}
		\label{fig:wrench_tracking_policy_vel}
	\end{subfigure}
	\caption{Wrench tracking experiment with enabled safety framework. All the quantities are shown in the body $x$ axis. The vertical dashed line indicates the time instant in which the cart is pulled away.}
	\label{fig:wrench_tracking_policy}
\end{figure}

In this experiment, we show that the safety layer allows us to safely perform wrench tracking in the presence of a disturbance due to variations in the environment.
\rwEditor{For the interaction wrench measurement we use a Bota Systems \ac{FT} sensor\footnote{\url{www.botasys.com}} with a sampling rate of $\SI{800}{\hertz}$. The raw data is filtered with a second order Butterworth filter with cutoff frequency $\SI{3}{\radian\per\second}$.}
An overview of the experiment is in Fig.~\ref{fig:interaction_exp_teaser}. %
\rwNine{As we now require wrench tracking along the $x$ body direction, the selection matrix in \eqref{selected_control} is chosen as $\selectionMat=\blkdiag{1, \mat{O}_{5\times5}}$.}
We want to apply a linear pushing force ranging from $\SI{3}{\newton}$ to $\SI{5}{\newton}$ on a flat interaction surface.
The drone's interaction tool is first positioned in contact with the cart using the pose controller.
Then, the pushing force is commanded along the drone's body $x$ axis, which is placed normal to the surface.
The results of the experiments are shown in Fig.~\ref{fig:wrench_tracking_policy}.
\rwNine{Initially, when the force reference is commanded, the \ac{LLE} estimation in Fig. \ref{fig:wrench_tracking_policy_lle} converges to a nominal value of \SI{-0.2}{\per\second} given by the closed loop wrench tracking dynamics.
	This is different from the nominal convergence rate of \SI{-0.5}{\per\second} for the pose tracking dynamics.
	The nominal convergence rates for the two dynamics can be computed with \eqref{eq:lle_nominal}.
	As the commanded force varies in steps in Fig. \ref{fig:wrench_tracking_policy_force}, the \ac{LLE} estimation is slightly excited while still keeping negative values since the force is successfully regulated.
}
At around $\SI{30}{\second}$, the cart is pulled away making force tracking not possible anymore, and the aerial vehicle accelerates in the pushing direction.
As the force dynamics no longer converge, the relative \ac{LLE} consistently increases to positive values, engaging the safety layer and enforcing the platform to dissipate energy and stop, as shown in Fig.~\ref{fig:wrench_tracking_policy_power} and Fig.~\ref{fig:wrench_tracking_policy_vel}. It only takes $\SI{0.2}{\second}$ for the \ac{LLE} to raise over zero and start dissipating.
\rwNine{The difference between the nominal \ac{WTC} command $\wrenchCommand$ and the safety filtered command $\wrenchCommandAugment$ is highlighted in Fig.~\ref{fig:wrench_tracking_policy_command}.
	The filtered command follows the nominal one as long as the force dynamics converge.
	When the interaction surface is pulled away, the filtered command forces the system to stop even if the nominal action would keep pushing forward.
	Also, while the \ac{WTC} is enabled along certain body axes, the position controller with relative safety layer is still enabled along the other axes. This makes the system robust with respect to other disturbances that can arise during interaction, like aerodynamic wall effects from nearby surfaces.}

\rwThree{
	\subsection{Damping coefficient effect}
	To show the effect of the damping gain $\Klamb$ on the safety layer performance, the same experiment presented in sec. \ref{wrench_tracking_subsection} has been repeated multiple times with different gains in the Raisim simulation environment \cite{raisim}.
	\begin{figure}[!tb]
		\centering
		\begin{subfigure}{0.5\columnwidth}
			\includegraphics[width=0.98\columnwidth]{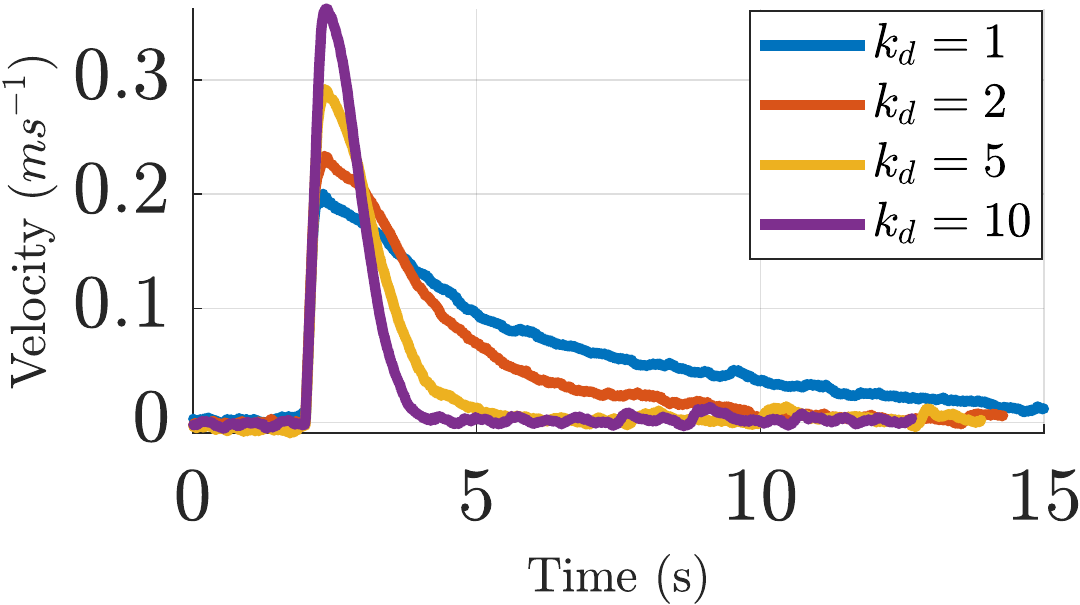}
			\caption{Linear velocity.}
			\label{fig:damping_vel}
		\end{subfigure}%
		\begin{subfigure}{0.5\columnwidth}
			\includegraphics[width=0.98\columnwidth]{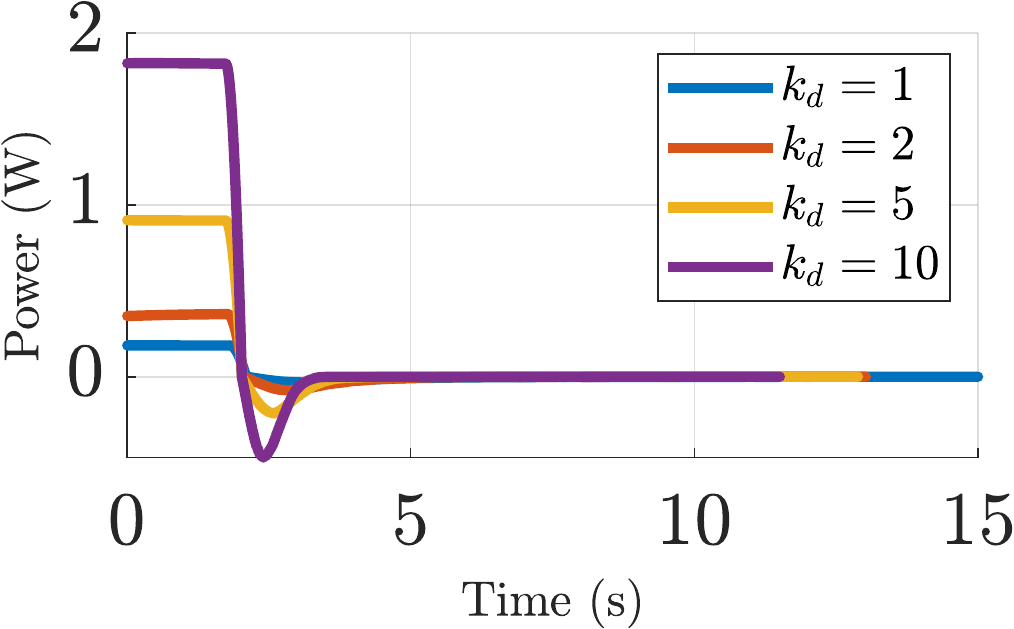}
			\caption{Linear power limitation.}
			\label{fig:damping_power}
		\end{subfigure}
		\caption{Simulated wrench tracking experiment with enabled safety framework for different values of the power damping gain $\Klamb$.}
		\label{fig:damping_raisim}
	\end{figure}
	As shown in Fig. \ref{fig:damping_vel}, increasing $\Klamb$ causes the system to dissipate its kinetic energy faster and stop sooner.
	This is given by more strict limits on the power dissipation, shown in Fig. \ref{fig:damping_power}.
	Also, the higher the value of $\Klamb$, the more power is allowed into the system when the \ac{LLE} is negative and no dissipation is required.
}
\section{Conclusions}
This work presented a safety layer for an \ac{OMAV} under pose tracking or wrench tracking controllers.
The method is based on power flow regulation and applicable to other fully actuated mechanical systems, since it relies only on the description of their closed loop dynamics.
Moreover, we imposed the power flow adaptive law with \acp{CBF} and proved it experimentally on the \ac{OMAV} platform.
Future works will investigate the proposed safety framework also in \ac{HRI} tasks.

\bibliographystyle{IEEEtran}
\bibliography{./lyap_ral_2022_bib}

\end{document}